\documentclass[a4paper,fleqn]{cas-sc}

\usepackage[numbers]{natbib}
\usepackage{placeins}
\usepackage{bbm}
\usepackage{xspace}
\newcommand{\eg}{e.g.\xspace}
\begin{document}
\let\WriteBookmarks\relax

\shorttitle{ORACLE-CT: Anatomy-Aware Support Pooling for CT Classification}
\shortauthors{Lavsen Dahal et~al.}

\title [mode = title]{ORACLE-CT: Anatomy-Aware Support Pooling for CT Classification}    

\author[1,2]{Lavsen Dahal}[orcid=0000-0002-8991-759X]
\cormark[1]
\ead{lavsen.dahal@duke.edu}

\credit{Conceptualization of this study, Methodology, Software, Writing - Original draft preparation}

\author[1,3]{Yubraj Bhandari}[orcid=0009-0004-7279-4097]
\ead{yubraj.bhandari@duke.edu}
\credit{Data curation, Methodology}

\author[4]{Geoffrey Rubin}[orcid=0000-0002-3820-2500]
\ead{grubin@arizona.edu}

\credit{Supervision, Resources}

\author[1,2]{Joseph Y. Lo}[orcid=0000-0002-9540-5072]
\cormark[2]
\ead{joseph.lo@duke.edu}
\credit{Supervision, Funding, Resources}

\affiliation[1]{organization={Center for Virtual Imaging Trials, RAI Labs, Department of Radiology, Duke University},
                city={Durham},
               postcode={27708}, 
                state={North Carolina},
                country={USA}}

\affiliation[2]{organization={Electrical and Computer Engineering, Pratt School of Engineering, Duke University},
                city={Durham},
               postcode={27708}, 
                state={North Carolina},
                country={USA}}      

\affiliation[3]{organization={Department of Mathematics, Trinity College of Arts \& Sciences, Duke University},
                city={Durham},
               postcode={27708}, 
                state={North Carolina},
                country={USA}}   
                
\affiliation[4]{organization={Department of Radiology and Imaging Sciences, University of Arizona College of Medicine},
                postcodesep={}, 
                city={Tucson},
                postcode={85004}, 
                state={Arizona},
                country={USA}}

\cortext[cor1]{Corresponding author}
\cortext[cor2]{Principal corresponding author}


\begin{abstract}
Abdominal CT disease classification is challenging because each examination is a large three-dimensional study with multiple possible findings, while diagnostic evidence is often localized to specific organs or anatomical compartments. Most study-level CT classifiers aggregate spatial encoder features using anatomy-agnostic operators such as global average pooling or unrestricted attention, allowing evidence for each label to be drawn from the full feature lattice. This creates a mismatch between anatomy-structured disease evidence and anatomy-agnostic evidence aggregation. We propose \textbf{ORACLE--CT}, an encoder-agnostic anatomy-aware aggregation framework that uses multi-organ segmentation to define label-specific anatomical supports and restricts attention pooling to those supports. ORACLE--CT supports single-organ, multi-organ union, comparative, localized, and global support strategies, enabling disease-specific evidence aggregation.

We evaluate ORACLE--CT across three representation families: a generic pretrained Vision Transformer (DINOv3), a native 3D convolutional network (I3D--ResNet-121), and a radiology-native foundation encoder (Pillar--0). Models are trained end-to-end on MERLIN and evaluated internally and under frozen external transfer to Duke--Abdomen and AMOS. Compared with global average pooling, support-masked pooling improved MERLIN macro-AUROC/AUPRC from 0.838/0.638 to 0.858/0.676 for DINOv3 and from 0.829/0.617 to 0.848/0.659 for I3D--ResNet-121. On the harmonized 10-label external evaluation, support-masked pooling also improved transfer for DINOv3 on Duke--Abdomen (AUROC/AUPRC: 0.802/0.628 to 0.835/0.683) and AMOS (0.742/0.313 to 0.762/0.350), with similar gains for I3D--ResNet-121. For Pillar--0, most improvement arose from learned attention itself, with smaller additional gains from support-masked pooling. These results show that enforcing anatomical support at pooling time improves discrimination and external robustness while preserving an auditable link between disease predictions and anatomical evidence.
\end{abstract}



\maketitle

\section{Introduction}
\label{sec:intro}
Abdominal CT is one of the most common and information-rich imaging examinations in clinical practice, but its increasing use has also contributed substantially to radiology workload~\cite{oecd_ct_exams_indicator, smith2025projected,vosshenrich2021quantifying,pourvaziri2022imaging}. This has motivated growing interest in automated interpretation of abdominal CT examinations, where clinically relevant findings may be numerous, co-occurring, and distributed across a large three-dimensional volume~\cite{shui2025large,agrawal2025pillar}. A central challenge is that diagnostically relevant evidence is not uniformly distributed across the scan, but is often anatomically anchored to specific organs, compartments, or inter-organ relationships~\cite{tushar2021classification,silverman2019bosniak,pickhardt2024post}. Effective CT interpretation therefore requires not only strong local image features, but also a principled mechanism for aggregating spatial evidence into examination-level decisions.

In this work, we focus on one scalable formulation of automated abdominal CT interpretation: multi-label classification of clinically relevant findings from abdominal CT examinations. Recent progress has made this formulation increasingly feasible at scale. Volumetric encoders and radiology foundation models now provide strong spatial representations~\cite{radford2021learning,wald2025comprehensive,blankemeier2024merlin,lin2024ct}. In parallel, natural language processing and large language model--assisted report extraction have made it increasingly practical to derive examination-level labels from radiology reports across large imaging datasets~\cite{mukherjee2023feasibility,reichenpfader2024scoping,le2024performance,garcia2025evaluating}. At the same time, robust multi-organ segmentation has become practical across large CT datasets, providing coarse organ and compartment maps that can serve as anatomical supports~\cite{wasserthal2023totalsegmentator,dahal2025xcat,Siemens_syngovia_accessed2025,Synopsys_Simpleware_AutoSeg_accessed2025}. Together, these developments create an opportunity for anatomically structured CT classification: strong spatial features, scalable examination-level supervision, and anatomical support maps are now available within a common modeling pipeline.

However, the availability of these components does not by itself specify how local image evidence should be converted into examination-level predictions. Modern CT classification pipelines typically pair a spatial encoder with an aggregation operator that summarizes local features for each target finding~\cite{draelos2021machine,liu2025does}. In practice, this step is often anatomy-agnostic: convolutional models commonly use global average pooling~\cite{lin2013network} over the final feature map, while Transformer-based models may use a learned class token that summarizes the full token sequence~\cite{dosovitskiy2020image}, or unrestricted attention over the entire spatial feature lattice. In such designs, evidence for a given finding can be drawn from the full volume unless anatomical support is explicitly imposed. 

Such generic aggregation can be mismatched to abdominal CT, where the relevant anatomical support varies across findings. Renal cysts and hydronephrosis are primarily kidney-centered~\cite{silverman2019bosniak}; hepatic steatosis is commonly assessed using hepatic attenuation and, in some protocols, liver--spleen attenuation ratio~\cite{haghshomar2024diagnostic,pickhardt2024post}; and ascites or metastatic disease may require evidence from broader but still anatomically constrained regions~\cite{cho2020peritoneal,meyers1973distribution}. Thus, aggregation is not simply a generic pooling step, but a central modeling choice. This motivates anatomy-aware classification strategies that guide evidence aggregation using organ and compartment supports.

\begin{figure*}
  \centering
  \includegraphics[width=\textwidth]{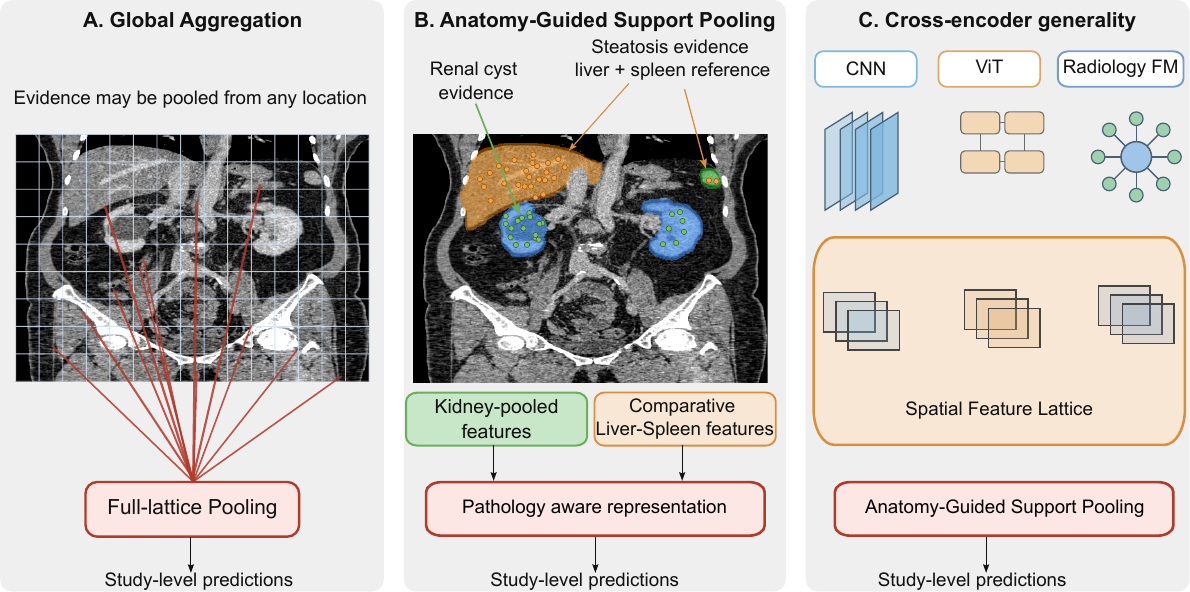}
\caption{\textbf{Conceptual overview of ORACLE--CT.}
(A) Standard study-level classification commonly pools over the full spatial feature lattice, potentially mixing pathology-relevant and irrelevant anatomy.
(B) ORACLE--CT assigns each pathology to anatomically appropriate support and pools features only within that support, producing pathology-aware representations such as kidney-pooled features for renal cyst and liver--spleen pooled features for hepatic steatosis.
(C) This anatomy-guided support-pooling principle is encoder-agnostic and can be instantiated on top of CNN, Vision Transformer, and radiology foundation model feature lattices.}
\label{fig:oracle_concept}
\end{figure*}

We address this problem with \textbf{ORACLE--CT}, an encoder-agnostic aggregation framework for study-level CT classification. ORACLE--CT keeps the full-volume encoder unchanged, but changes where the classifier is allowed to pool spatial evidence for each finding. Each label is assigned to a clinically motivated anatomical support. Pooling is then performed within the assigned support rather than indiscriminately over the entire feature lattice. In this way, anatomical structure is introduced at the representation-formation stage, while preserving compatibility with diverse CT encoders.

To isolate the role of anatomical support, we evaluate ORACLE--CT under a controlled head-comparison protocol against anatomy-agnostic aggregation baselines. For each encoder family, the backbone, preprocessing, and optimization protocol are held fixed while only the aggregation mechanism is varied, separating the effect of label-specific support restriction from generic changes in pooling capacity. We test this principle across three encoder families: a natural-image pretrained Vision Transformer, a native 3D CNN, and a radiology-native foundation encoder. Figure~\ref{fig:oracle_concept} illustrates the contrast between anatomy-agnostic global aggregation and pathology-specific support pooling.

This work makes four main contributions:
\begin{enumerate}
    \item We propose a modular support-masked aggregation head that can be attached to heterogeneous CT encoders and trained end-to-end without cropping the input volume or modifying the encoder architecture.
    
    \item We define a clinically motivated support taxonomy for abdominal CT findings, spanning single-organ, multi-organ union, comparative, localized, and global supports, enabling organ-specific, relational, and diffuse findings to be modeled within a unified framework.
    
    \item We introduce a controlled aggregation comparison that separates anatomy-agnostic pooling from anatomy-aware support pooling while keeping the encoder family, preprocessing, and optimization protocol fixed within each backbone.
    
    \item We validate ORACLE--CT on MERLIN and two external datasets, Duke--Abdomen and AMOS, showing improved internal performance and frozen external transfer across generic transformer, native 3D CNN, and radiology-native foundation-model representations.
\end{enumerate}

\section{Related Work}

\textbf{Supervised study-level CT classification.}
Large-scale supervised CT classification studies have shown that multi-label disease prediction from volumetric CT is feasible, while also highlighting persistent challenges in weak supervision, missing labels, class imbalance, calibration, and external validation~\cite{draelos2021machine,tushar2021classification}. 
Prior work has also incorporated anatomical structure by dividing CT interpretation across body regions, for example using separate models or classifiers for different anatomical regions or organ systems~\cite{tushar2021classification}. 
These approaches demonstrate the value of anatomical information for CT disease classification, but they primarily use anatomy to partition the input space or specialize prediction models. 
They leave open a complementary question: when a shared full-volume encoder is used, how should spatial features be aggregated into label-specific study-level predictions?
This question is especially relevant for scalable multi-label systems, where a single encoder pass may need to support many findings whose evidence arises from different anatomical supports.

\textbf{CT report supervision, vision-language models, and foundation encoders.}
Recent CT representation learning methods have focused on building stronger image encoders by pretraining on large-scale CT datasets, often using paired CT volumes and radiology reports through contrastive, generative, or multimodal objectives~\cite{hamamci2024developing,lin2024ct,blankemeier2024merlin,shui2025large,agrawal2025pillar,li2026greenrfm,lee2025unified,husegmentation,beeche2025pan,wald2025comprehensive}. 
These approaches have enabled CT vision-language models and radiology foundation encoders that can support downstream classification through zero-shot prompting, linear probing, or end-to-end fine-tuning. Their central strength is improved spatial representation learning and transfer across CT tasks. 
However, once a spatial feature representation has been extracted, study-level classification still requires an aggregation mechanism that determines how local features are summarized into finding-specific predictions. 

\textbf{Aggregation, multiple instance learning, and anatomical priors.}
In large-scale CT classification, labels are often available for the entire examination, while the relevant image evidence may occupy only a subset of the volume. The model must therefore summarize many local token or voxel features into one prediction for each finding. This setting is closely related to multiple instance learning (MIL), where a bag of spatial instances is mapped to a global label under weak supervision~\cite{dietterich1997solving,ilse2018attention,zaheer2017deep}.

Learned attention pooling provides a flexible way to weight spatial instances before forming an examination-level representation~\cite{ilse2018attention,bahdanau2014neural,luong2015effective,xu2015show}. However, when applied without anatomical constraints, attention pooling still learns weights over the full feature lattice and does not specify which regions are clinically appropriate for a given prediction. This motivates aggregation strategies that retain full-volume encoding while incorporating anatomical support into the pooling stage.

\section{Method: ORACLE--CT}
\label{sec:method}

\begin{figure*}
  \centering
  \includegraphics[width=\textwidth]{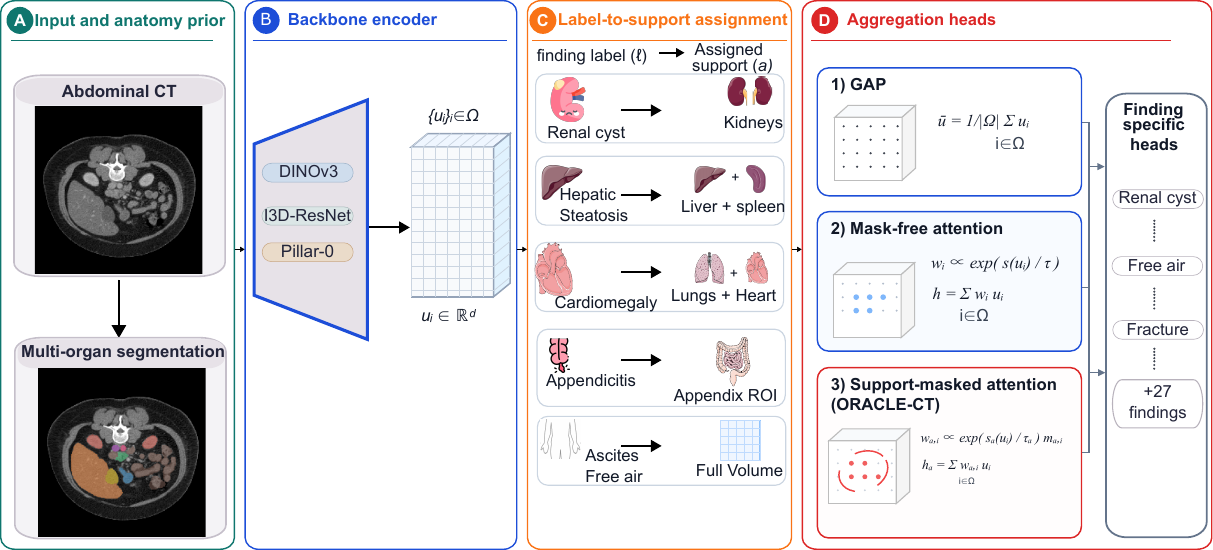}
    \caption{\textbf{Overview of ORACLE--CT anatomy-aware support pooling.}
    ORACLE--CT modifies the aggregation stage of multi-label CT classification while leaving the encoder architecture unchanged.
    Each backbone is trained separately, but within a given backbone the same full-volume encoder provides features for all labels.
    (A) An abdominal CT volume is paired with a multi-organ segmentation, which provides coarse anatomical supports.
    (B) A pretrained CT encoder maps the full volume to a spatial feature lattice \(\{u_i\}_{i\in\Omega}\). ORACLE--CT is encoder-agnostic and is evaluated separately with DINOv3, I3D--ResNet-121, and Pillar--0.
    (C) Each finding label is assigned to a predefined anatomical support, such as a single organ, a comparative multi-organ support, a union support, a localized region, or the full volume when global context is appropriate.
    (D) Three aggregation heads are compared under the same training protocol: global average pooling uniformly averages the full lattice; mask-free attention pooling learns content-weighted aggregation over all locations; and support-masked attention pooling normalizes attention only within the label-specific anatomical support.
    The resulting support-aware representations are passed to finding-specific classification heads for study-level multi-label prediction.}
    \label{fig:oracle_overview}
\end{figure*}

Figure~\ref{fig:oracle_overview} summarizes ORACLE--CT. We consider supervised multi-label CT classification, where a single CT volume is used to predict a set of finding labels. At a high level, ORACLE--CT decomposes this task into three components:
(i) a pretrained encoder that maps the input volume to a dense spatial feature lattice,
(ii) segmentation-derived anatomical supports assigned to finding labels, and
(iii) an aggregation head that pools evidence within the assigned support before classification.

This decomposition isolates the effect of anatomy-aware aggregation by holding the encoder family and training protocol fixed within each backbone, and yields support-restricted pooling weights that can be qualitatively audited.

\subsection{Problem Setup}
\label{sec:problem_setup}

A CT volume is represented as \(X\in\mathbb{R}^{D\times H\times W}\). 
We consider a supervised multi-label classification problem with \(L\) finding labels. 
Ground-truth targets are \(y\in\{0,1,-1\}^{L}\), where \(-1\) denotes a missing or unknown label. 
An encoder \(E_{\theta}\) extracts spatial features from the volume, and an aggregation pathway \(A_{\phi}\) maps these features to label logits \(z\in\mathbb{R}^{L}\):
\[
z \;=\; A_{\phi}\!\big(E_{\theta}(X)\big), 
\qquad
\hat{\mathbf p}\;=\;\mathrm{sigmoid}(z)\in[0,1]^L.
\]
Here, \(A_{\phi}\) denotes the full aggregation pathway, including label-to-support assignment, support-aware pooling, and finding-specific classification heads. Our focus is how this pathway should use segmentation-derived anatomical supports to convert spatial encoder features into anatomically meaningful multi-label predictions.

\subsection{Encoders as Feature Lattices}
\label{sec:backbones}

As illustrated in Figure~\ref{fig:oracle_overview}B, ORACLE--CT treats each backbone as a feature-lattice generator and applies the same downstream aggregation logic to the resulting spatial representation. Let \(\Omega\) denote the index set of the encoder's final spatial feature lattice, and let \(\{u_i\}_{i\in\Omega}\), with \(u_i\in\mathbb{R}^d\), denote the local features consumed by the aggregation head. The construction of \(\Omega\) depends on the encoder substrate, but all pooling modes operate only on the resulting spatial features. This abstraction makes ORACLE--CT compatible with heterogeneous CT encoders while isolating the contribution of the aggregation head.

In this work, we validate ORACLE--CT across three representative backbone families: 
\emph{(i)} a pseudo-3D token lattice produced by applying a 2D transformer to axial tri-slice inputs (DINOv3~\cite{simeoni2025dinov3}; after dropping non-spatial tokens), 
\emph{(ii)} a voxel lattice produced by a native 3D CNN (I3D--ResNet-121)~\cite{he2016deep,carreira2017quo}, and 
\emph{(iii)} a spatial activation lattice produced by a radiology-native foundation encoder (Pillar--0)~\cite{agrawal2025pillar} evaluated under its native preprocessing and recommended input geometry. 
We write \(i\in\Omega\) for a lattice location, whether it corresponds to a token or a voxel-like spatial cell, and flatten coordinates for brevity.

DINOv3 provides a strong generic pretrained transformer encoder without radiology-specific pretraining. Because DINOv3 is a 2D vision transformer, we used a 2.5D tri-slice implementation: each axial center slice \(c\) was represented as a three-channel image using adjacent slices \((c-1,c,c+1)\), resized to \(224\times224\), ImageNet-normalized, and processed independently by the pretrained encoder. After removing the class token and DINOv3 register tokens, the remaining spatial patch tokens were reassembled across sampled axial positions to form a pseudo-3D token lattice. Support masks were aligned using the same axial centers: the corresponding mask center slices were extracted, resized with nearest-neighbor interpolation to the DINOv3 input resolution, and downsampled to the patch-token grid for support-masked softmax pooling. 

I3D--ResNet-121 represents a native 3D convolutional architecture for volumetric feature extraction and corresponds to the visual encoder family used in the MERLIN vision-language model~\cite{blankemeier2024merlin}. We include this backbone to test whether ORACLE--CT improves downstream multi-label classification when built on a MERLIN-style volumetric image encoder. For this backbone, the CT volume was processed directly as a 3D tensor, and the final convolutional feature map was treated as a voxel lattice, with each spatial cell corresponding to one instance \(u_i\). Segmentation-derived support masks were resampled with nearest-neighbor interpolation to the same feature-map resolution before pooling.

Pillar--0 represents a radiology-native foundation encoder pretrained at scale on CT data. Pillar--0 provides both a global pooled feature vector and a spatial activation map. The GAP baseline used the pooled output, whereas mask-free attention pooling and support-masked pooling operated on the spatial activation map after resampling segmentation-derived supports to the activation lattice. Thus, despite differences in architecture and pretraining, all three encoders are interfaced through the same ORACLE--CT abstraction: a spatial feature lattice followed by an aggregation head. This setup lets us test whether the value of explicit anatomy-aware aggregation depends on encoder inductive bias and pretraining domain.

\subsection{Organ Groups, Anatomical Supports, and Label Anchoring}
\label{sec:masks}

We introduce an explicit anatomy prior through segmentation-derived masks and a fixed label-to-support assignment. Figure~\ref{fig:oracle_overview}A shows the segmentation-derived anatomy prior, while Figure~\ref{fig:oracle_overview}D summarizes the representative support types used across disease labels.

\paragraph{Organ groups.}
Raw segmentation classes \(\mathcal{C}\) are merged into a smaller set of clinically meaningful \emph{organ groups} \(\mathcal{O}\)
(\eg left/right kidneys \(\rightarrow\) \texttt{kidneys}).
For each raw class \(c \in \mathcal{C}\), let \(S_c \in \{0,1\}^{D\times H\times W}\) denote its binary mask.
Let \(\pi:\mathcal{C}\to\mathcal{O}\) map raw classes to groups and define the group mask
\[
M_o \;=\; \bigvee_{c:\,\pi(c)=o} S_c \in \{0,1\}^{D\times H\times W},
\qquad o\in\mathcal{O}.
\]
We dilate \(M_o\) by \(r_o\) mm in metric space to absorb boundary uncertainty, then resample the mask to the encoder lattice \(\Omega\) to obtain indicators \(m_{o,i}\in\{0,1\}\) and support
\[
\Omega_o=\{\,i\in\Omega:\, m_{o,i}=1\,\}.
\]

\paragraph{Label anchoring and anatomical supports.}
Rather than mapping each label directly to a single organ, ORACLE--CT maps each disease label to a predefined anatomical support. Let \(\mathcal{A}\) denote the set of support definitions, where each support \(a\in\mathcal{A}\) may correspond to a single organ, a union of organs, a comparative multi-organ support, a localized anatomical region, or the full volume. Each label \(\ell\in\{1,\dots,L\}\) is assigned to one support through a fixed map
\[
\kappa:\{1,\dots,L\}\to\mathcal{A}.
\]
This induces support-specific label sets
\[
\mathcal{L}_a=\{\ell:\kappa(\ell)=a\}.
\]
For each support \(a\), we construct a binary lattice mask \(m_{a,i}\in\{0,1\}\) when the support is mask-defined. Labels assigned to the same support share the corresponding pooled representation, while predictions remain disease-specific through separate output units in the support-specific classifier head. Global labels use the full lattice without an anatomical mask. For mask-defined supports, we denote the corresponding lattice index set by
\[
\Omega_a=\{i\in\Omega:m_{a,i}=1\}.
\]

\subsection{From Feature Lattices to ORACLE Heads}
\label{sec:agg-overview}

Given \(\{u_i\}_{i\in\Omega}\), we compare three encoder-compatible aggregation heads in a staged ladder
(Figure~\ref{fig:oracle_overview}B) to isolate the effect of content weighting and anatomy priors:
\textbf{(i) GAP} (Eq.~\ref{eq:gap-head}),
\textbf{(ii) mask-free attention pooling} (Eqs.~\ref{eq:attn-pool-weights}--\ref{eq:attn-pool-head}), and
\textbf{(iii) support-masked attention pooling} (Eqs.~\ref{eq:masked-alpha}--\ref{eq:masked-pool-classify}).

\subsubsection{Mask-free Heads}
\label{sec:heads-maskfree}

\paragraph{Global Average Pooling (GAP).}
The simplest baseline summarizes the entire feature lattice uniformly:
\begin{equation}
  h \;=\; \frac{1}{|\Omega|}\sum_{i\in\Omega} u_i \;\in\; \mathbb{R}^{d},
  \qquad
  z \;=\; W^{\mathrm{global}} h + b^{\mathrm{global}} \;\in\; \mathbb{R}^{L}.
  \label{eq:gap-head}
\end{equation}
Here \(W^{\mathrm{global}}\!\in\!\mathbb{R}^{L\times d}\) and \(b^{\mathrm{global}}\!\in\!\mathbb{R}^{L}\).
GAP is feature-only, \(\mathcal{O}(|\Omega|)\), and serves as our supervised baseline.

\paragraph{Mask-free attention pooling (unary baseline).}
To move beyond uniform pooling, we assign a learned weight to each lattice location using a lightweight \emph{unary} scorer, followed by softmax normalization over the full lattice:
\begin{equation}
  \alpha_i \;=\; s_{\mathrm{pool}}(u_i),
  \qquad
  w_i \;=\; \frac{\exp(\alpha_i/\tau)}{\sum_{j\in\Omega}\exp(\alpha_j/\tau)},
  \quad \tau>0,
  \label{eq:attn-pool-weights}
\end{equation}
\begin{equation}
  h \;=\; \sum_{i\in\Omega} w_i\,u_i \in \mathbb{R}^{d},
  \qquad
  z \;=\; W^{\mathrm{global}} h + b^{\mathrm{global}} \in \mathbb{R}^{L}.
  \label{eq:attn-pool-head}
\end{equation}
The scorer \(s_{\mathrm{pool}}:\mathbb{R}^d\!\to\!\mathbb{R}\) is a small per-location map
(Linear\((d{\to}1)\) on tokens or \(1{\times}1{\times}1\) Conv3D on voxels).
The weights satisfy \(w_i\!\ge\!0\) and \(\sum_{i\in\Omega} w_i\!=\!1\).
Compared with GAP, mask-free attention pooling learns a content-weighted mixture while remaining mask-free and encoder-compatible.

\noindent\emph{Note:}
This is attention \emph{pooling} via unary scoring and softmax normalization, not Transformer-style Q/K/V feature mixing~\cite{vaswani2017attention,wang2018non}.
It is included specifically to decouple content-weighted pooling from the anatomical masking used by ORACLE--CT.

\subsubsection{Support-masked Head}
\label{sec:heads-organ}

\paragraph{Support-masked attention pooling.}
For each anatomical support \(a\in\mathcal{A}\), we apply a support-specific unary scorer
\(s_a:\mathbb{R}^d\!\to\!\mathbb{R}\) and temperature \(\tau_a>0\):
\begin{equation}
  \alpha_{a,i} = s_a(u_i).
  \label{eq:masked-alpha}
\end{equation}
For mask-defined supports, attention is normalized only over locations inside the support using a masked softmax:
\begin{equation}
  w_{a,i}
  =
  \frac{\exp(\alpha_{a,i}/\tau_a)\,m_{a,i}}
       {\sum_{j\in\Omega}\exp(\alpha_{a,j}/\tau_a)\,m_{a,j}+\varepsilon},
  \qquad \varepsilon>0.
  \label{eq:masked-softmax}
\end{equation}
The support-specific pooled representation and logits are then
\begin{equation}
  h_a = \sum_{i\in\Omega} w_{a,i}u_i,
  \qquad
  z_a = W^{(a)}h_a+b^{(a)}\in\mathbb{R}^{|\mathcal{L}_a|}.
  \label{eq:masked-pool-classify}
\end{equation}
Here \(W^{(a)}\) and \(b^{(a)}\) are classifier parameters for the disease labels assigned to support \(a\).

For global supports, no anatomical mask is applied and the corresponding labels use the mask-free global attention head. If a mask-defined support \(\Omega_a\) is empty after resampling to the encoder lattice, we instead use the mask-free global attention representation for labels assigned to that support. This fallback is applied before computing the support-specific pooled representation and prevents degenerate empty-support outputs when small organs or thin structures vanish after downsampling.

\paragraph{Composite and comparative supports.}
For union supports, the support mask \(m_{a,i}\) is formed by the union of the component organ masks and pooling proceeds as in Eq.~\ref{eq:masked-pool-classify}. For comparative supports, component organs are pooled separately and then concatenated before classification. For example, for hepatic steatosis,
\[
h_{\mathrm{steatosis}}
=
\left[
h_{\mathrm{liver}};\,
h_{\mathrm{spleen}}
\right],
\qquad
z_{\mathrm{steatosis}}
=
W_{\mathrm{steatosis}} h_{\mathrm{steatosis}} + b_{\mathrm{steatosis}}.
\]
This allows the classifier to use cross-organ information while preserving organ-specific pooling.

Support-masked attention differs from both GAP and mask-free attention in one crucial respect: the normalization domain is anatomically constrained. As a result, the pooled representation for each support-anchored disease label is formed only from locations within the assigned segmented support, encouraging study-level predictions to rely on anatomically appropriate support regions rather than unrestricted global context.


\subsection{Training Objective}
\label{sec:train-objective}

\paragraph{Loss with missing finding labels.}
We optimize binary cross-entropy only over findings with available study-level labels. 
Let \(m_\ell=\mathbbm{1}[y_\ell\neq -1]\) denote the label-availability mask for finding \(\ell\), where \(y_\ell=-1\) indicates that the finding label is missing or uncertain. For available labels, \(y_\ell\in\{0,1\}\), and the per-study loss is
\[
\mathcal{L}
=
\frac{1}{\sum_{\ell=1}^{L} m_\ell}
\sum_{\ell=1}^{L}
m_\ell\,
\mathrm{BCEWithLogits}(z_\ell,y_\ell).
\]
The BCE term was weighted using label-specific positive weights estimated from the MERLIN training split, with positive-weight ratios clipped at 10.

\section{Experimental Design, datasets, and Evaluation Protocol}
\label{sec:expdesign}

\begin{table*}[t]
\centering
\caption{\textbf{Abdominal CT datasets and standard input geometry.}
Counts reflect patient-disjoint splits. MERLIN served as the internal development dataset, while Duke--Abdomen and AMOS were used only for external evaluation. For frozen transfer, calibration parameters and operating thresholds were selected on MERLIN validation and applied unchanged to the external datasets. For the standard DINOv3 and I3D--ResNet-121 experiments, CT volumes were clipped to $[-1000,1000]$ HU, min--max normalized to $[0,1]$, and resampled to the common geometry shown below. Pillar--0 was evaluated using its native 11-window input representation and input geometry, as described in Table~\ref{tab:train-settings}. Duke--Abdomen shares 27 labels with MERLIN, and AMOS shares a 10-label subset used for three-way evaluation.}
\label{tab:data-brief}
\begin{tabular}{lrrrrcc}
\toprule
Dataset & Train & Val & Test & Labels & Resampled Spacing (mm) & Shape \\
\midrule
MERLIN (internal) & 15{,}175 & 5{,}018 & 5{,}082 & 30 & $3{\times}1.5{\times}1.5$ & $160{\times}224{\times}224$ \\
Duke--Abdomen (external) & \multicolumn{2}{c}{---} & 2{,}000 & 27 & $3{\times}1.5{\times}1.5$ & $160{\times}224{\times}224$ \\
AMOS (external) & \multicolumn{2}{c}{---} & 1{,}107 & 10 & $3{\times}1.5{\times}1.5$ & $160{\times}224{\times}224$ \\
\bottomrule
\end{tabular}
\end{table*}

\subsection{Datasets}
\label{sec:data}

Table~\ref{tab:data-brief} summarizes the abdominal CT datasets, split sizes, label spaces, and standard input geometry used for the DINOv3 and I3D--ResNet-121 experiments. Unless otherwise noted, intensities were clipped to $[-1000,1000]$ HU and min--max normalized to $[0,1]$.

\paragraph{MERLIN (internal development dataset).}
MERLIN is an abdomen-focused CT dataset with 30 study-level labels annotated as positive, negative, or missing/unknown~\cite{blankemeier2024merlin}. We used the released patient-disjoint train/validation/test splits and treated MERLIN as the sole internal development dataset for model training, model selection, calibration, and threshold selection.

\paragraph{Duke--Abdomen (external clinical dataset).}
We additionally evaluated on a private abdominal CT dataset from Duke University Hospitals (\(N{=}2{,}000\)), with 27 study-level labels overlapping MERLIN. Duke--Abdomen was used only for frozen external evaluation, with no training, recalibration, or threshold tuning performed on this dataset. The Duke--Abdomen dataset was assembled from de-identified archival abdominal CT examinations with associated radiology reports under an approved Duke University Health System IRB protocol.

\paragraph{AMOS (external public dataset).}
We further evaluated on the public AMOS abdominal CT dataset (\(N{=}1{,}107\)), harmonized to a 10-label subset shared with MERLIN and Duke--Abdomen for three-way comparison. 

\subsection{Reference Labels and Label Harmonization}
\label{sec:labels}

MERLIN labels were used as released with the dataset. For Duke--Abdomen, study-level reference labels were derived from radiology reports using two independent LLM-assisted NLP pipelines based on MedGemma~\cite{sellergren2025medgemma} and Qwen~\cite{yang2025qwen3}, implemented within the RATE framework released with Pillar--0~\cite{agrawal2025pillar}. Each pipeline extracted report findings and mapped them to the MERLIN study-level label space. Concordant positive or negative assignments were accepted as the final reference label, whereas disagreements or insufficient report evidence for a given finding were marked as missing/unknown and excluded for that finding.

For AMOS, we used the publicly released labels from the Foundation Models for General CT Image Diagnosis challenge~\cite{codabench_general_ct_diagnosis}; no AMOS labels were modified or re-derived. Because label availability differed across datasets, evaluation was performed on the dataset-specific evaluable label set: 30 labels in MERLIN, 27 labels overlapping MERLIN in Duke--Abdomen, and 10 AMOS labels that could be unambiguously harmonized with MERLIN findings set. Macro-averaged performance was therefore reported within each dataset's evaluable label set, while the shared 10-label subset was used for direct three-way comparison across MERLIN, Duke--Abdomen, and AMOS.

\subsection{Segmentation and Findings-Specific Support Construction}
\label{sec:grouping}

\paragraph{Segmentation backbone.}
We obtained multi-organ segmentations using TotalSegmentator~\cite{wasserthal2023totalsegmentator} as anatomical coordinate systems for defining label-specific support regions used during support-masked attention pooling.

\paragraph{Findings-specific anatomical supports.}
For each abdominal finding, we defined a support region according to the expected spatial organization of its imaging evidence rather than assigning all labels to a single fixed organ grouping scheme. Unless otherwise noted, mask-defined supports were dilated by 3~mm in metric space before pooling to reduce sensitivity to segmentation boundary uncertainty. The same label-to-support mapping was applied across MERLIN, Duke--Abdomen, and AMOS. No dataset-specific remapping or support redesign was performed for external evaluation; only the evaluable label subset differed across datasets. Table~\ref{tab:oracle_roi_strategy} summarizes the anatomical support definitions used for the abdominal findings.

\begin{table*}[t]
\centering
\scriptsize
\setlength{\tabcolsep}{5pt}
\renewcommand{\arraystretch}{1.12}
\caption{\textbf{Disease-specific anatomical supports used by ORACLE--CT.}
Each target finding is assigned a label-dependent support strategy defined from multi-organ segmentation. Unless otherwise noted, mask-defined supports were dilated by 3~mm before pooling; bone masks were not dilated, cardiomegaly used a larger-context cardiothoracic support, comparative supports used separate organ-wise pooling followed by feature concatenation, and global labels used no anatomical mask. For supports requiring contextual proxies, support definitions were fixed before model training and applied unchanged across datasets. Pleural-context support was constructed by dilating the lung masks to include adjacent pleural space. As direct appendix mask was unavailable, the appendicitis support was approximated using an appendix-focused lower-right bowel/cecal region.}
\label{tab:oracle_roi_strategy}
\begin{tabular}{p{0.23\linewidth} p{0.12\linewidth} p{0.27\linewidth} p{0.28\linewidth}}
\toprule
\textbf{Disease label(s)} & \textbf{Strategy} & \textbf{Support construction} & \textbf{Pooling behavior} \\
\midrule

Hepatomegaly 
& Single-organ 
& Liver support 
& Attention restricted to the \texttt{liver} mask \\

Splenomegaly 
& Single-organ 
& Spleen support 
& Attention restricted to the \texttt{spleen} mask \\

Abdominal aortic aneurysm  
& Single-organ 
& Aortic support
& Attention restricted to the \texttt{aorta} mask \\

Gallstones 
& Single-organ 
& Gallbladder support 
& Attention restricted to the \texttt{gallbladder} mask \\

Hydronephrosis 
& Single-organ 
& Bilateral kidney support 
& Union of \texttt{kidney\_left} and \texttt{kidney\_right} masks \\

Renal hypodensities 
& Single-organ 
& Bilateral kidney support 
& Union of \texttt{kidney\_left} and \texttt{kidney\_right} masks \\

Atherosclerosis 
& Single-organ 
& Aortic support 
& Attention restricted to the \texttt{aorta} mask \\

Coronary calcification 
& Single-organ 
& Cardiac support 
& Attention restricted to the \texttt{heart} mask \\

Atelectasis 
& Single-organ 
& Lung support 
& Union of lung-lobe masks \\

Osteopenia, Fracture 
& Single-organ 
& Skeletal support 
& Union of available bone masks; no dilation applied \\

Pancreatic atrophy 
& Single-organ 
& Pancreatic support 
& Attention restricted to the \texttt{pancreas} mask \\

Prostatomegaly 
& Single-organ 
& Prostate support 
& Attention restricted to the \texttt{prostate} mask \\

\midrule

Biliary ductal dilation 
& Union 
& Hepatobiliary-pancreatic support 
& Union of \texttt{liver}, \texttt{gallbladder}, and \texttt{pancreas} masks \\

Cardiomegaly 
& Union 
& Cardio-thoracic support 
& Union of \texttt{heart} and lung masks with larger-context dilation \\

Bowel obstruction, Submucosal edema 
& Union 
& Bowel support 
& Union of \texttt{small\_bowel} and \texttt{colon} masks \\

Pleural effusion 
& Union 
& Lung + pleural-context support 
& Union of lung masks with pleural-space context/proxy region \\

Hiatal hernia 
& Union 
& Upper GI / thoracoabdominal junction support 
& Union of \texttt{stomach} and lung masks \\

Renal cyst 
& Union 
& Kidney + cyst support 
& Union of bilateral kidney masks with available kidney-cyst masks \\

Surgically absent gallbladder 
& Union 
& Gallbladder fossa support 
& Union of \texttt{gallbladder} and \texttt{liver} masks \\

Aortic valve calcification 
& Union 
& Aortic-root / cardiac support 
& Union of \texttt{heart} and \texttt{aorta} masks \\

\midrule

Hepatic steatosis 
& Comparative 
& Liver vs spleen comparative support 
& Liver and spleen pooled separately, then concatenated before classification \\

\midrule

Appendicitis 
& Localized  
& Appendix-focused subregion within bowel support 
& Localized appendix proxy derived from the bowel/cecal region \\

\midrule

Thrombosis, Free air, Ascites, Anasarca, Metastatic disease, Lymphadenopathy 
& Global 
& Full-volume support 
& No anatomical mask applied; prediction uses the mask-free full-lattice attention head \\

\bottomrule
\end{tabular}
\end{table*}

\subsection{Preprocessing and Augmentation}
\label{sec:preprocess}

For the standard DINOv3 and I3D--ResNet-121 experiments, CT volumes were clipped to $[-1000,1000]$ HU, min--max normalized to $[0,1]$, and resampled to the voxel spacing and input shape listed in Table~\ref{tab:data-brief}. Pillar--0 was evaluated using its native 11-window input representation and input geometry, as shown in Table~\ref{tab:train-settings}. As Pillar-0 retains its native preprocessing pipeline, its results are reported throughout for contextual cross-encoder comparison rather than strict preprocessing-matched comparison. During MERLIN training, we used mild 3D augmentation consisting of small affine perturbations (in-plane rotation, translation, and scaling), gamma intensity jitter, and low-variance additive noise. Spatial flips were disabled along all axes to preserve anatomical laterality and organ correspondence (e.g., spleen-left versus liver-right).

\subsection{Training Protocol}
\label{sec:train-protocol}

External datasets were never used for optimization, model selection, recalibration, or threshold tuning. To isolate the effect of study-level aggregation, all backbones were fine-tuned end-to-end on MERLIN under a harmonized training protocol, with only backbone-specific adjustments to learning rate, batch size, and memory-saving settings when required by input geometry.

We optimized a multi-label binary cross-entropy objective using AdamW with cosine learning-rate decay, mixed-precision training, and gradient clipping. For the standard ORACLE--CT experiments with DINOv3 and I3D--ResNet-121, models were trained for 20 epochs with a base learning rate of \(3\times10^{-4}\), weight decay \(10^{-4}\), batch size 10, and 5 warm-up epochs. Grouped learning rates were used so that the classification head learned faster than the backbone, with head learning-rate scale \(=3.0\). Validation was performed every epoch, and the checkpoint with the highest validation macro-AUROC was retained for final testing.

Training used class-balanced positive weights derived from the MERLIN training split, with positive-weight ratios clipped at 10. Labels marked missing or unknown (\(-1\)) were excluded from both loss computation and evaluation.

All reported DINOv3 and I3D--ResNet-121 results were obtained by end-to-end fine-tuning rather than frozen-feature evaluation. For the Pillar--0 experiments, we retained the model's native 11-window abdominal CT input representation and used a lower base learning rate (\(1\times10^{-4}\)), shorter warm-up of 2 epochs, batch size 1 with gradient accumulation over 4 steps, and gradient checkpointing to accommodate the substantially larger \(384^3\) input volume. 
\begin{table*}[t]
\centering
\small
\setlength{\tabcolsep}{5pt}
\renewcommand{\arraystretch}{1.08}
\caption{\textbf{Backbone-specific training settings.}
All models were fine-tuned end-to-end on MERLIN and selected by validation macro-AUROC.}
\label{tab:train-settings}
\begin{tabular}{lccccc}
\toprule
\textbf{Backbone} & \textbf{Input} & \textbf{Batch} & \textbf{LR} & \textbf{Warm-up} & \textbf{Notes} \\
\midrule
DINOv3 & $160{\times}224{\times}224$ & 10 & $3\times10^{-4}$ & 5 & AMP, AdamW, cosine decay \\
I3D--ResNet-121 & $160{\times}224{\times}224$ & 10 & $3\times10^{-4}$ & 5 & AMP, AdamW, cosine decay \\
Pillar--0 & $11{\times}384^3$ & 1 & $1\times10^{-4}$ & 2 & grad accum.\ 4, checkpointing \\
\bottomrule
\end{tabular}
\end{table*}

\subsection{Evaluation Metrics, Calibration, and Operating Points}
\label{sec:metrics}
We report threshold-free discrimination using macro-AUROC and macro-AUPRC, computed as unweighted averages over labels with available ground truth in each dataset. Labels marked missing or unknown were excluded from metric computation.

For threshold-dependent metrics, we applied temperature scaling on MERLIN validation and then selected per-model, per-label operating points by maximizing F1 score on the same split. F1 and balanced accuracy were both computed using these F1-optimized thresholds.

\subsection{Label Prevalence and Distribution Shift}
\label{sec:prevalence}

We analyzed label prevalence to characterize class imbalance in the internal dataset and findings-frequency shift across external abdomen datasets. MERLIN exhibited a long-tailed label distribution, with substantial variation in prevalence across findings and several relatively rare labels (Supplementary Figure S2). External evaluation was therefore performed not only under institutional shift, but also under nontrivial prevalence mismatch.

\begin{figure*}
  \centering
  \includegraphics[width=\textwidth]{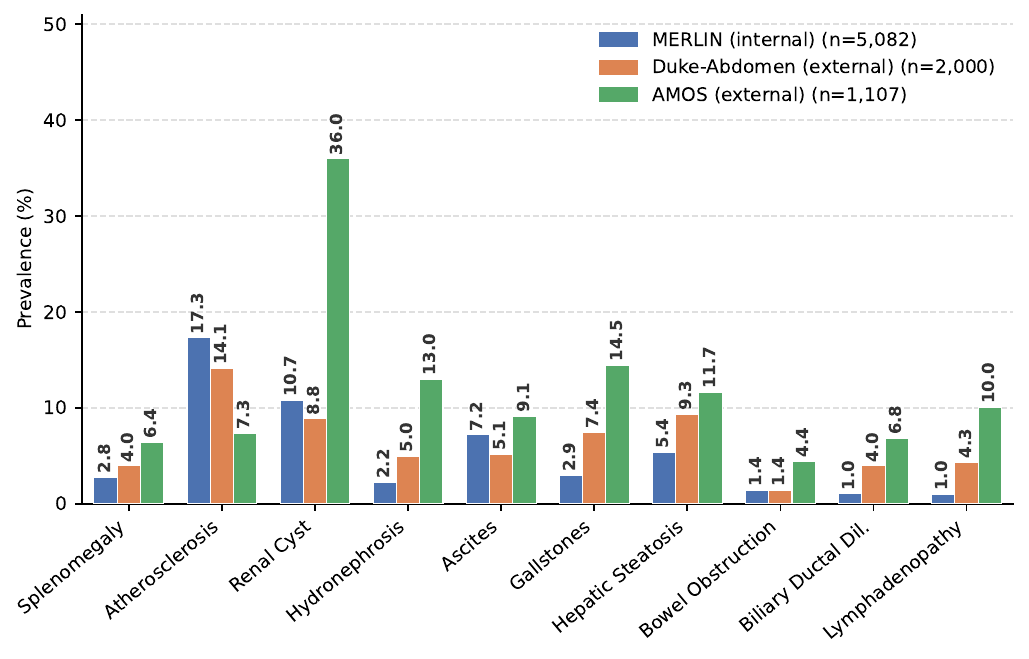}
\caption{\textbf{Prevalence of the harmonized label subset across test datasets.}
Per-label prevalence (\%) for the labels shared across MERLIN (internal; \(n=5{,}082\)), Duke--Abdomen (external; \(n=2{,}000\)), and AMOS (external; \(n=1{,}107\)).
Marked prevalence differences remain across datasets despite label harmonization, indicating substantial dataset-level variation in case mix.}
  \label{fig:prev_abd_harmonized}
\end{figure*}

Figure~\ref{fig:prev_abd_harmonized} compares per-label prevalence across MERLIN, Duke--Abdomen, and AMOS for the harmonized 10-label subset used for three-way evaluation. Several findings showed marked prevalence differences between datasets, indicating that external transfer was assessed under genuine cross-dataset case-mix shift.

\section{Results}

\subsection{Support-masked pooling improves internal classification across encoder families}
\label{sec:results_internal}

We first evaluated whether anatomy-aware support pooling improves study-level classification on the MERLIN internal test split. This experiment isolates the effect of the aggregation head by comparing three encoder-compatible pooling modes: GAP, mask-free attention pooling, and support-masked pooling while keeping the backbone family and end-to-end training protocol fixed.

Table~\ref{tab:merlin_modes_all_backbones} shows an encoder-dependent pattern. For DINOv3 and I3D--ResNet-121, support-masked pooling achieved the strongest overall performance. For DINOv3, macro-AUROC improved from 0.8380 with GAP to 0.8576 with support-masked pooling, while macro-AUPRC increased from 0.6382 to 0.6763. For I3D--ResNet-121, macro-AUROC increased from 0.8288 to 0.8482 and macro-AUPRC from 0.6169 to 0.6587. These gains indicate that, for a generic pretrained transformer and a native 3D convolutional encoder, study-level performance is limited not only by local feature quality but also by how spatial evidence is aggregated.

Pillar--0 exhibited a different profile. Replacing GAP with mask-free attention pooling produced the dominant improvement, increasing macro-AUROC from 0.8003 to 0.8562 and macro-AUPRC from 0.5980 to 0.6925. Support-masked pooling yielded a similar macro-AUROC of 0.8578, with a smaller and metric-dependent incremental benefit over mask-free attention.

\begin{figure*}[!htbp]
    \centering
    \includegraphics[width=\linewidth]{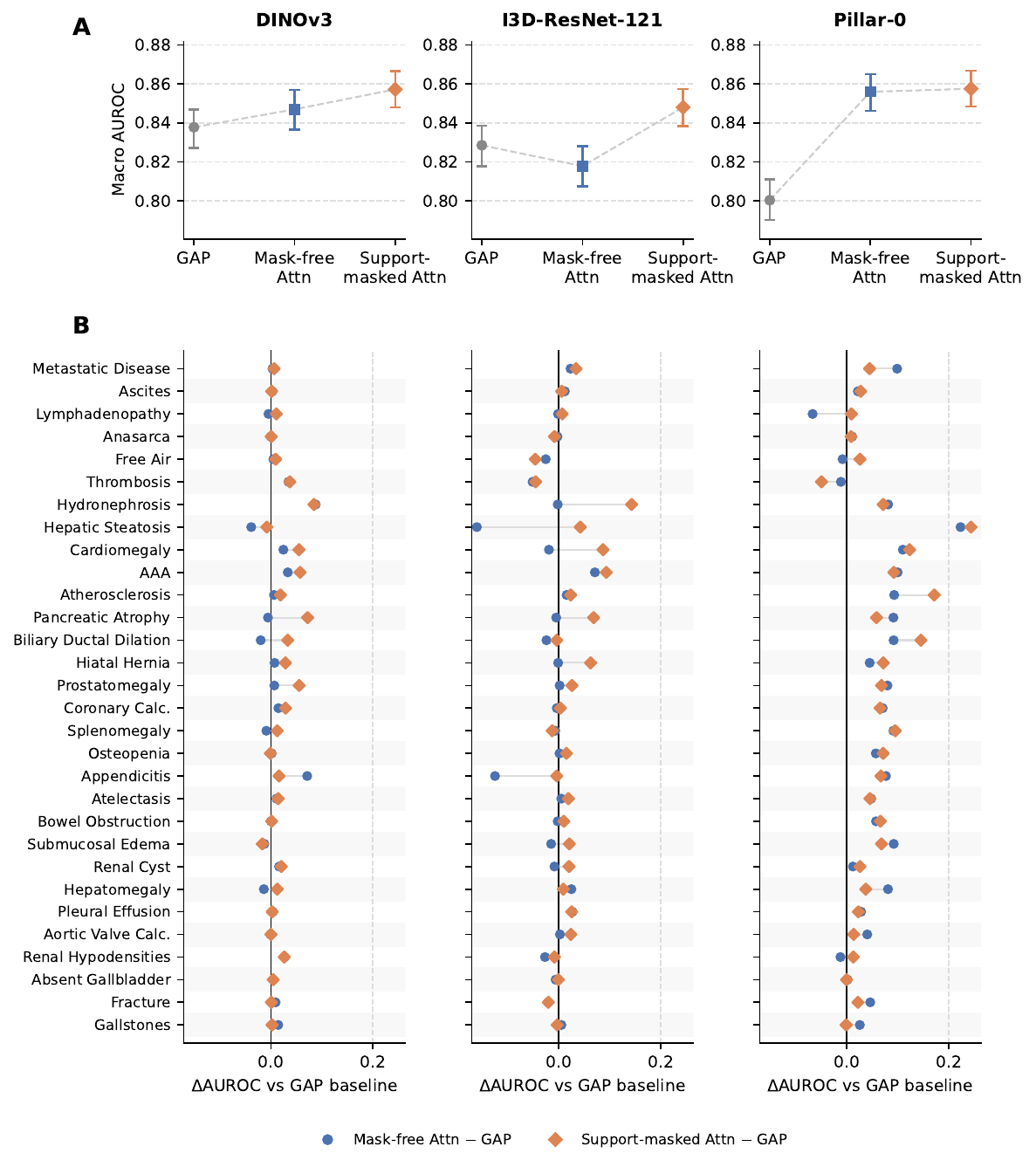}
 \caption{\textbf{ORACLE-CT: Encoder-dependent internal test results on MERLIN (n=5,082).}
(A) Macro-averaged AUROC on the MERLIN test set across three pooling modes for each encoder backbone. Error bars denote 95\% bootstrap confidence intervals (1,000 bootstrap replicates).
(B) Per-disease $\Delta$AUROC relative to the GAP baseline for mask-free attention pooling (blue circles) and support-masked attention pooling (orange diamonds). Diseases are ordered by the average support-masked $\Delta$AUROC gain over GAP across the three backbones (ascending from bottom to top).
For DINOv3 and I3D--ResNet-121, support-masked pooling yields broader positive disease-level shifts and the strongest dataset-level AUROC, whereas for Pillar--0 the dominant gain arises from replacing GAP with learned attention, with support masking providing a smaller additional benefit.}
    \label{fig:merlin_internal_disease_gains}
\end{figure*}

Figure~\ref{fig:merlin_internal_disease_gains} provides a disease-level view of the same pattern. For DINOv3 and I3D--ResNet-121, support-masked pooling produced broader positive \(\Delta\)AUROC shifts relative to GAP than mask-free attention pooling. In contrast, Pillar--0 showed a smaller additional gain from support masking after learned attention was introduced. Together, these results suggest that explicit anatomical support restriction is most beneficial when anatomical structure is not already strongly represented by the backbone. Full 95\% bootstrap confidence intervals for all metrics are provided in Supplementary Table S1.

\begin{table*}[t]
\centering
\setlength{\tabcolsep}{6pt}
\small
\caption{\textbf{Internal abdomen mode ablations on MERLIN across encoder families.}
Mode ablations on the MERLIN test split for a generic pretrained token backbone (DINOv3), a native 3D CNN (I3D--ResNet-121), and a radiology foundation encoder (Pillar--0).
We compare uniform global average pooling, mask-free attention pooling, and support-masked pooling.
Metrics are macro-averaged over the full MERLIN label set.
Pillar–0 results are reported for contextual cross-encoder comparison (see \ref{sec:preprocess}). Point estimates are shown for readability. Full 95\% bootstrap confidence intervals for all metrics are provided in Supplementary Table S1.}
\label{tab:merlin_modes_all_backbones}
\begin{tabular}{llcccc}
\toprule
\textbf{Backbone} & \textbf{Mode} & \textbf{AUROC} & \textbf{AUPRC}& \textbf{F1}  & \textbf{BA}  \\
\midrule
\multirow{3}{*}{DINOv3}
& GAP baseline              & 0.8380 & 0.6382 & 0.6356 & 0.7302 \\
& Mask-free attn pooling    & 0.8473 & 0.6617 & 0.6432 & 0.7263 \\
& Support-masked pooling      & \textbf{0.8576} & \textbf{0.6763} & \textbf{0.6596} & \textbf{0.7471} \\

\midrule
\multirow{3}{*}{I3D--ResNet-121}
& GAP baseline              & 0.8288 & 0.6169 & 0.6108 & 0.7124 \\
& Mask-free attn pooling    & 0.8181 & 0.6086 & 0.6048 & 0.7104 \\
& Support-masked pooling      & \textbf{0.8482} & \textbf{0.6587} & \textbf{0.6424} & \textbf{0.7427} \\

\midrule
\multirow{3}{*}{Pillar--0}
& GAP baseline              & 0.8003 & 0.5980 & 0.5972 & 0.6923 \\
& Mask-free attn pooling    & 0.8562 & \textbf{0.6925} & 0.6571 & \textbf{0.7437} \\
& Support-masked pooling      & \textbf{0.8578} & 0.6884 & \textbf{0.6598} & 0.7424 \\
\bottomrule
\end{tabular}
\end{table*}

\subsection{Support-masked pooling improves frozen external transfer}
\label{sec:results_external}

We next evaluated whether the internal gains persisted under frozen external transfer. All models were trained on MERLIN, and model selection, calibration parameters, and operating thresholds were selected on MERLIN validation and transferred unchanged to Duke--Abdomen and AMOS.

On the harmonized 10-label subset shared across MERLIN, Duke--Abdomen, and AMOS, support-masked pooling achieved the highest macro-AUROC across all three encoder families (Table~\ref{tab:abdomen_harmonized10}). For DINOv3, support-masked pooling improved AUROC from 0.8022 to 0.8353 on Duke--Abdomen and from 0.7424 to 0.7616 on AMOS relative to GAP. AUPRC increased from 0.6276 to 0.6828 on Duke--Abdomen and from 0.3125 to 0.3502 on AMOS. For I3D--ResNet-121, AUROC improved from 0.7751 to 0.8197 on Duke--Abdomen and from 0.7135 to 0.7345 on AMOS, with corresponding AUPRC gains from 0.5903 to 0.6461 and from 0.2936 to 0.3273.

For Pillar–0, the encoder-dependent pattern observed internally persisted under transfer: on Duke–Abdomen, mask-free attention improved AUROC from 0.7818 to 0.8488, and support-masked pooling further increased AUROC to 0.8538. On AMOS, the corresponding values were 0.7137, 0.7570, and 0.7714.

\begin{figure*}[!t]
    \centering
    \includegraphics[width=\textwidth]{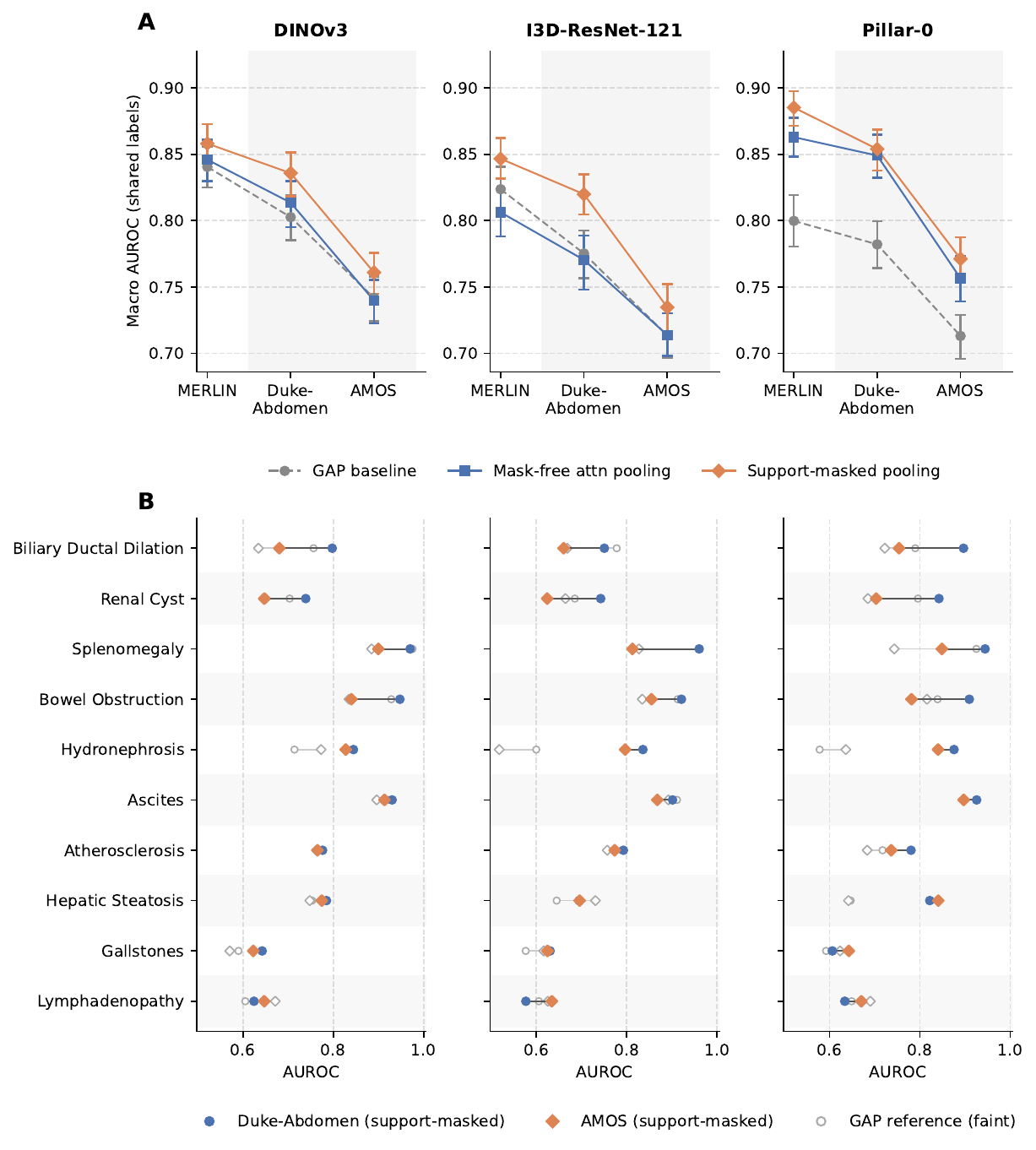}
    \caption{\textbf{External generalization on the harmonized 10-label subset.}
    (A) Macro AUROC across MERLIN, Duke--Abdomen, and AMOS for GAP, mask-free attention pooling, and support-masked pooling across DINOv3, I3D--ResNet-121, and Pillar--0. Error bars denote 95\% bootstrap confidence intervals.
    (B) Disease-level decomposition of the Duke-to-AMOS shift for the support-masked model on the 10 harmonized labels. Blue circles denote Duke AUROC and orange diamonds denote AMOS AUROC, with horizontal black segments connecting the same disease across datasets; longer leftward shifts indicate larger degradation on AMOS. Hollow gray markers show the corresponding GAP baseline values as a faint reference. Diseases are ordered by average Duke-to-AMOS AUROC change under support-masked pooling across backbones.
    While support-masked pooling remains the strongest dataset-level transfer setting in Panel A, Panel B shows that the AMOS performance drop is not uniform across labels, but is disproportionately driven by a subset of harmonized findings.}
    \label{fig:external_generalization_h10}
\end{figure*}

Figure~\ref{fig:external_generalization_h10} summarizes the harmonized external-transfer results. At the dataset level, support-masked pooling remained the strongest AUROC setting across backbones, while all methods showed a larger performance drop on AMOS than on Duke--Abdomen. At the disease level, the Duke-to-AMOS degradation was not uniform across labels; instead, a subset of findings accounted for a disproportionate share of the macro-AUROC reduction. Thus, the lower AMOS performance reflects not only broad dataset shift but also disease-specific differences in detectability, label definition, or case mix. Full 95\% bootstrap confidence intervals for the harmonized 10-label evaluation are reported in Supplementary Table S2.

\begin{table*}[t]
\centering
\setlength{\tabcolsep}{4pt}
\small
\caption{\textbf{External abdomen generalization on the harmonized 10-label subset across encoder families.}
Performance on the 10 findings shared across MERLIN, Duke--Abdomen, and AMOS. We compare the GAP baseline, mask-free attention pooling, and support-masked pooling.
Metrics are macro-averaged over the harmonized label set.
Pillar--0 is included for contextual comparison (see \ref{sec:preprocess}). Point estimates are shown for readability. Full 95\% bootstrap confidence intervals for all metrics are provided in Supplementary Table S2.}

\label{tab:abdomen_harmonized10}
\begin{tabular}{llcccccc}
\toprule
& & \multicolumn{2}{c}{\textbf{MERLIN}} & \multicolumn{2}{c}{\textbf{Duke}} & \multicolumn{2}{c}{\textbf{AMOS}} \\
\cmidrule(lr){3-4}\cmidrule(lr){5-6}\cmidrule(lr){7-8}
\textbf{Backbone} & \textbf{Mode}
& \textbf{AUROC} & \textbf{AUPRC}
& \textbf{AUROC} & \textbf{AUPRC}
& \textbf{AUROC} & \textbf{AUPRC} \\
\midrule
\multirow{3}{*}{DINOv3}
& GAP baseline           & 0.8405 & 0.6727 & 0.8022 & 0.6276 & 0.7424 & 0.3125 \\
& Mask-free attn pooling & 0.8461 & 0.6963 & 0.8131 & 0.6494 & 0.7401 & 0.3214 \\
& Support-masked pooling   & \textbf{0.8583} & \textbf{0.7042} & \textbf{0.8353} & \textbf{0.6828} & \textbf{0.7616} & \textbf{0.3502} \\
\midrule
\multirow{3}{*}{I3D--ResNet-121}
& GAP baseline           & 0.8238 & 0.6411 & 0.7751 & 0.5903 & 0.7135 & 0.2936 \\
& Mask-free attn pooling & 0.8064 & 0.6370 & 0.7700 & 0.5740 & 0.7141 & 0.2905 \\
& Support-masked pooling   & \textbf{0.8468} & \textbf{0.6705} & \textbf{0.8197} & \textbf{0.6461} & \textbf{0.7345} & \textbf{0.3273} \\
\midrule
\multirow{3}{*}{Pillar--0}
& GAP baseline           & 0.7999 & 0.6517 & 0.7818 & 0.5781 & 0.7137 & 0.2850 \\
& Mask-free attn pooling & 0.8630 & \textbf{0.7458} & 0.8488 & \textbf{0.7034} & 0.7570 & 0.3928 \\
& Support-masked pooling   & \textbf{0.8854} & 0.7448 & \textbf{0.8538} & 0.7032 & \textbf{0.7714} & \textbf{0.4022} \\
\bottomrule
\end{tabular}
\end{table*}

\begin{table*}[t]
\centering
\setlength{\tabcolsep}{4pt}
\small
\caption{\textbf{MERLIN-to-Duke transfer on the shared 27-label Abdominal CT subset.}
Performance is reported on the 27 findings shared between MERLIN and Duke--Abdomen. Pillar--0 is included for contextual cross-encoder comparison  (see \ref{sec:preprocess}).  Full 95\% bootstrap confidence intervals for Duke--Abdomen are provided in Supplementary Table S3.}
\label{tab:abdomen_merlin_duke27}
\begin{tabular}{llcccccccc}
\toprule
& & \multicolumn{4}{c}{\textbf{MERLIN (27-label subset)}} & \multicolumn{4}{c}{\textbf{Duke--Abdomen (27-label subset)}} \\
\cmidrule(lr){3-6}\cmidrule(lr){7-10}
\textbf{Backbone} & \textbf{Mode}
& \textbf{AUROC} & \textbf{AUPRC} & \textbf{F1} & \textbf{BA}
& \textbf{AUROC} & \textbf{AUPRC} & \textbf{F1} & \textbf{BA} \\
\midrule
\multirow{3}{*}{DINOv3}
& GAP baseline           & 0.8368 & 0.6344 & 0.6345 & 0.7274 & 0.8212 & 0.6460 & 0.6194 & 0.7151 \\
& Mask-free attn pooling & 0.8466 & 0.6599 & 0.6441 & 0.7257 & 0.8312 & 0.6709 & 0.6134 & 0.7229 \\
& Support-masked pooling & \textbf{0.8582} & \textbf{0.6771} & \textbf{0.6610} & \textbf{0.7460} & \textbf{0.8483} & \textbf{0.6988} & \textbf{0.6335} & \textbf{0.7375} \\

\midrule
\multirow{3}{*}{I3D--ResNet-121}
& GAP baseline           & 0.8269 & 0.6126 & 0.6091 & 0.7098 & 0.7911 & 0.6143 & 0.5964 & 0.6875 \\
& Mask-free attn pooling & 0.8165 & 0.6043 & 0.6029 & 0.7086 & 0.7933 & 0.6083 & 0.5888 & 0.7037 \\
& Support-masked pooling & \textbf{0.8471} & \textbf{0.6561} & \textbf{0.6422} & \textbf{0.7417} & \textbf{0.8280} & \textbf{0.6622} & \textbf{0.6269} & \textbf{0.7346} \\

\midrule
\multirow{3}{*}{Pillar--0}
& GAP baseline           & 0.7968 & 0.5902 & 0.5911 & 0.6880 & 0.7750 & 0.5940 & 0.5500 & 0.6680 \\
& Mask-free attn pooling & 0.8544 & \textbf{0.6894} & 0.6530 & \textbf{0.7393} & 0.8280 & 0.6820 & \textbf{0.6240} & \textbf{0.7210} \\
& Support-masked pooling & \textbf{0.8572} & 0.6867 & \textbf{0.6564} & 0.7386 & \textbf{0.8330} & \textbf{0.6930} & 0.6170 & 0.7160 \\
\bottomrule
\end{tabular}
\end{table*}

\subsection{Disease-level Duke--Abdomen transfer gains are broadly distributed but support-dependent}
\label{sec:results_duke_heatmap}

\begin{figure*}[!t]
    \centering
     \includegraphics[width=\textwidth]{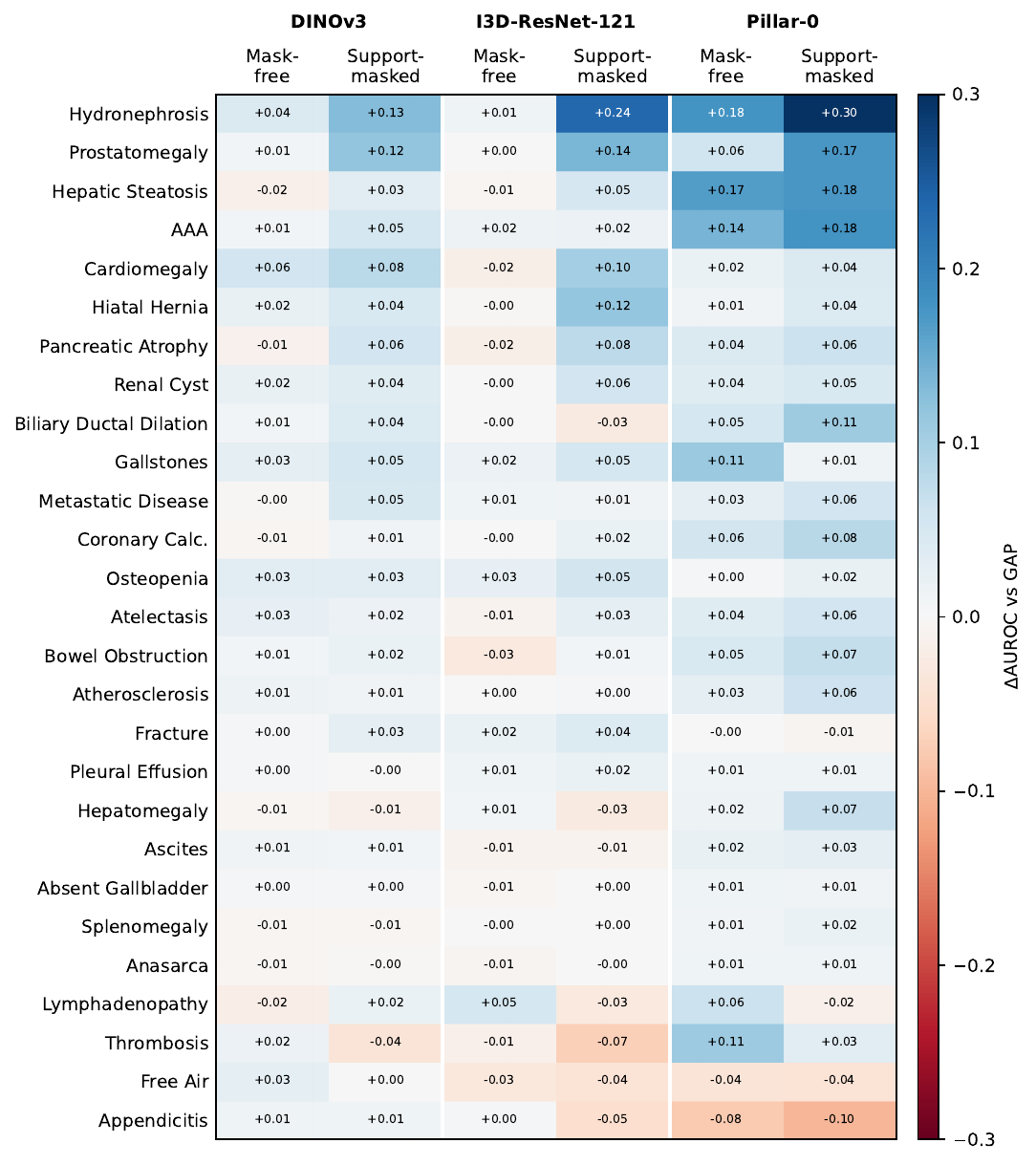}
    \caption{\textbf{Disease-level MERLIN-to-Duke transfer on the shared 27-label subset.}
    Heatmap of per-finding \(\Delta\)AUROC relative to GAP on Duke--Abdomen for mask-free attention pooling and support-masked pooling across DINOv3, I3D--ResNet-121, and Pillar--0. Blue indicates improvement over uniform full-lattice aggregation, red indicates lower AUROC than GAP, and values denote absolute AUROC differences. Rows are ordered by the average support-masked gain across encoders. Support-masked pooling produces broadly distributed gains for many anatomically localized or measurement-linked findings, whereas smaller or negative deltas occur for labels with sparse or variable evidence, low positive prevalence, incomplete annotation, or intentionally global support definitions where support-masked pooling reduces to full-lattice attention and therefore provides limited additional anatomical constraint. Full absolute per-disease AUROC values are provided in Supplementary Figure S1.}
    \label{fig:duke27_heatmap}
\end{figure*}

The harmonized 10-label analysis provides the cleanest three-dataset comparison, but it does not capture the full Duke--Abdomen label overlap. We therefore additionally evaluated MERLIN-to-Duke transfer on the broader 27-label subset shared between MERLIN and Duke--Abdomen. At the dataset level, support-masked pooling remained the strongest or near-strongest setting across encoder families (Table~\ref{tab:abdomen_merlin_duke27}); full 95\% bootstrap confidence intervals for this Duke shared-27 evaluation are provided in Supplementary Table S3. To determine whether these dataset-level gains were driven by a few labels or distributed across the broader Duke label space, we examined per-finding \(\Delta\)AUROC relative to GAP (Figure~\ref{fig:duke27_heatmap}). Full absolute per-disease AUROC values for the Duke shared-27 evaluation are provided in Supplementary Figure S1.

The heatmap shows that support-masked pooling produced broadly positive shifts across many Duke labels, particularly for anatomically localized or measurement-linked findings. Large and consistent gains were observed for hydronephrosis, prostatomegaly, hepatic steatosis, abdominal aortic aneurysm, renal cysts, biliary ductal dilation, cardiomegaly, and pancreatic atrophy. For example, hydronephrosis improved by \(+0.13\), \(+0.24\), and \(+0.30\) AUROC for DINOv3, I3D--ResNet-121, and Pillar--0, respectively. These disease-level patterns support the central hypothesis that label-specific anatomical support provides useful inductive bias when the spatial distribution of diagnostic evidence is well matched to the assigned support.

The heatmap also shows that support-masked pooling is not uniformly beneficial for every finding. Negative or minimal deltas were observed for a smaller subset of labels, including appendicitis, free air, thrombosis, and lymphadenopathy. These cases likely reflect different failure modes. For appendicitis, the localized appendix-focused support may be too narrow or sensitive to segmentation and downsampling errors. In contrast, thrombosis, free air, ascites, anasarca, metastatic disease, and lymphadenopathy were assigned global support because their evidence can be distributed, sparse, or anatomically variable; for these labels, support-masked pooling provides limited additional anatomical restriction beyond the full-lattice attention pathway. Several of these labels also have low positive prevalence and/or incomplete annotation in MERLIN, which can increase the variance of per-finding transfer estimates (Supplementary Figure S2). Thus, disease-level gains depend on both support suitability and the amount and quality of label supervision.

\subsection{Support-masked attention provides auditable anatomical evidence aggregation}
\label{sec:results_qualitative}

\begin{figure*}[!t]
    \centering
    \includegraphics[width=0.80\textwidth]{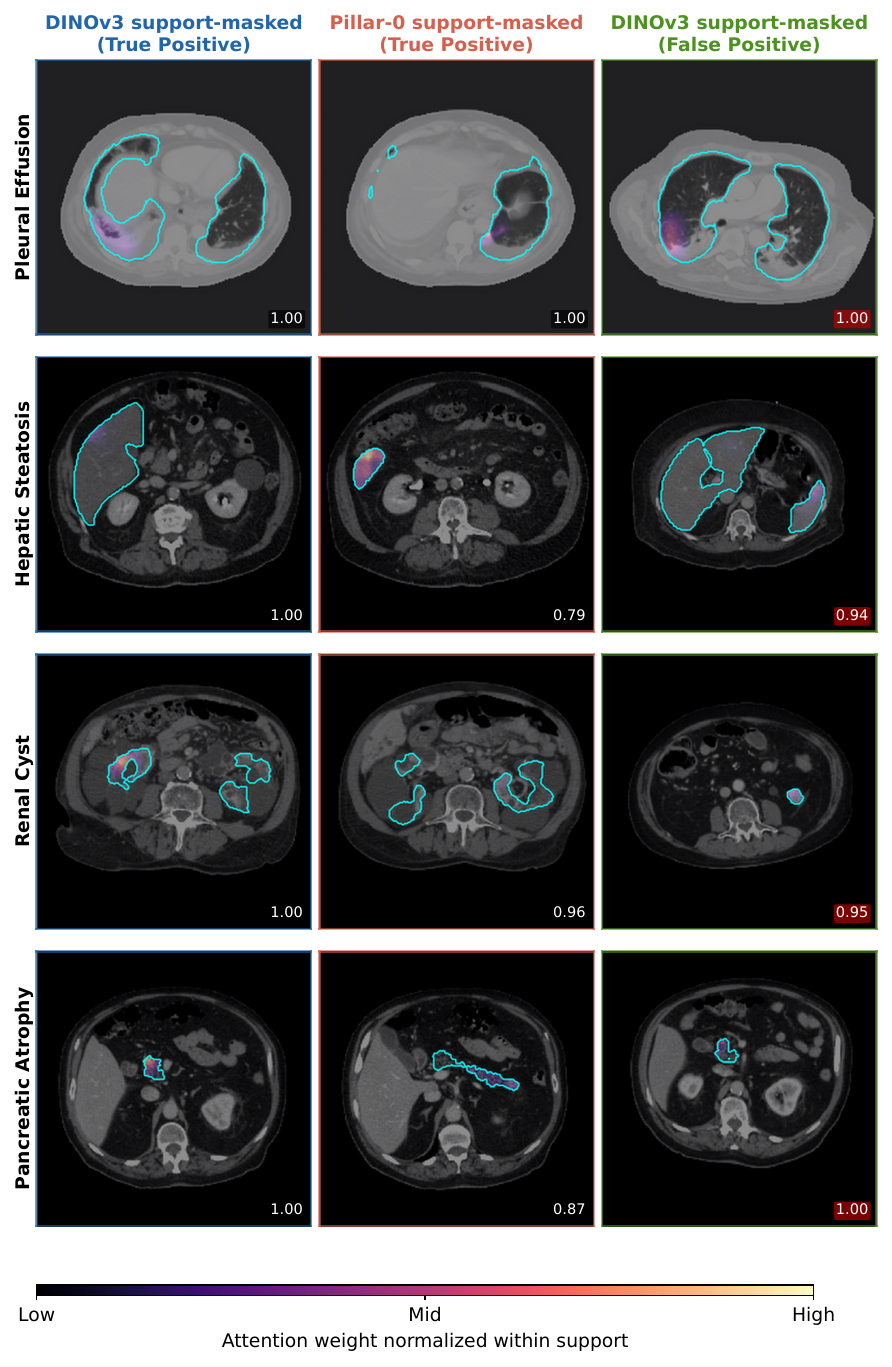}
    \caption{\textbf{Qualitative audit of support-masked attention.}
    Representative axial CT slices with anatomical support contours and support-masked attention overlays for pleural effusion, hepatic steatosis, renal cyst, and pancreatic atrophy. Columns show true-positive examples from DINOv3 support-masked and Pillar--0 support-masked models, and representative false-positive examples from DINOv3 support-masked models. Attention remains concentrated within the assigned anatomical support in both true-positive and false-positive examples, indicating that ORACLE--CT errors can occur despite anatomically plausible support localization. Score badges denote sigmoid output for the positive class.}
    \label{fig:qualitative_attention}
\end{figure*}

Finally, we examined whether support-masked attention weights could be used to audit the anatomical support from which study-level evidence was aggregated. Figure~\ref{fig:qualitative_attention} shows representative examples for pleural effusion, hepatic steatosis, renal cyst, and pancreatic atrophy, including true-positive cases from DINOv3 support-masked and Pillar--0 support-masked models, as well as representative false-positive cases from DINOv3 support-masked models.

Across true-positive examples, high-response regions remained confined to the predefined anatomical support for the corresponding label. For pleural effusion, attention was restricted to the lung and pleural-context support; for hepatic steatosis, responses were constrained to the comparative liver--spleen support; for renal cyst, attention remained within kidney-centered support; and for pancreatic atrophy, responses were confined to the pancreatic support. This pattern was observed for both the generic pretrained DINOv3 encoder and the radiology-native Pillar--0 encoder, suggesting that support-masked pooling enforces a similar anatomy-constrained evidence pathway despite substantial differences in backbone pretraining and feature representation.

The false-positive examples provide a complementary audit. Although the predicted labels were incorrect, the attention responses generally remained within the assigned anatomical support rather than drifting toward unrelated organs or background anatomy. These cases therefore suggest that some ORACLE--CT errors can occur despite anatomically plausible support localization, reflecting limitations of study-level supervision, label noise, visual ambiguity, or insufficient discriminative evidence within the correct support region rather than gross off-target aggregation.

These maps should be interpreted as qualitative audits of the aggregation support, not as definitive lesion-level explanations. Because voxel or lesion-level ground truth was unavailable for most findings, the overlays cannot prove that the classifier attended to the causal abnormality. Rather, they verify an important property of the proposed head: the evidence used for study-level pooling is constrained to the anatomical support assigned to each label.

\FloatBarrier

\section{Discussion}
\label{sec:discussion}

This study investigated whether multi-label abdominal CT classification benefits from making the aggregation stage explicitly anatomy-aware. Across internal and frozen external evaluation, replacing unrestricted full-lattice aggregation with label-specific support pooling improved performance most clearly for DINOv3 and I3D--ResNet-121. These gains indicate that, even with strong pretrained encoders, the way spatial evidence is summarized into finding-level predictions can substantially affect discrimination and transfer.

The benefit of support-masked pooling can be understood as a constraint on the spatial competition induced by attention pooling. Global average pooling compresses features without regard to anatomical compartment, while mask-free attention learns non-uniform weights but still normalizes over the full feature lattice. Support-masked pooling changes this normalization domain: evidence competes only within the anatomical support assigned to each label. This distinction was especially important for I3D--ResNet-121, where mask-free attention slightly underperformed GAP. A plausible explanation is that the native 3D CNN lattice is highly downsampled and locally smoothed, so an unconstrained unary attention head must learn both content weighting and anatomical localization from study-level labels alone. In that setting, attention may overemphasize non-specific high-response locations, producing a less stable summary than uniform averaging. Anatomical support restriction acts as a regularizer by converting attention from an unconstrained localization problem into a within-support evidence aggregation problem.

The Pillar--0 results refine this interpretation. For this radiology-native foundation encoder, most of the gain came from replacing GAP with learned attention, while explicit support masking produced a smaller and metric-dependent incremental benefit. This pattern suggests that radiology-specific pretraining may already encode part of the spatial and anatomical organization that ORACLE--CT introduces explicitly at pooling time. However, the masking benefit did not disappear, particularly for AUROC under external transfer, indicating that explicit anatomical support and radiology-native representation learning are partially complementary. In practice, support masking appears most valuable when anatomical priors are weak in the encoder, but it can still provide structure and auditability when stronger medical pretraining is available.

The disease-level analyses show that the value of anatomy-aware aggregation is support-dependent. Gains were largest for findings whose diagnostic evidence is anatomically localized. In contrast, smaller or negative deltas occurred for labels whose assigned support was either difficult to define precisely or intentionally global. Appendicitis may be sensitive to a narrow appendix-focused proxy and to segmentation or downsampling errors, whereas thrombosis, free air, ascites, anasarca, metastatic disease, and lymphadenopathy were assigned global support because their evidence can be sparse, distributed, or anatomically variable. For such labels, support-masked pooling provides limited additional restriction beyond full-lattice attention. Several of these findings also have low positive prevalence or incomplete annotation in MERLIN, which can make disease-level estimates less stable. 

The same analysis also highlights a limit of fixed support design. Not all findings are intrinsic to a single organ; some depend on a broader
reference frame. Hepatomegaly is the clearest example - liver size is interpreted relative to patient body habitus, and recent habitus-aware
abdominal anatomy generation work similarly models organ size as a residual relative to body scale rather than as an isolated
volume~\cite{bhandari2026abdomengen}. A strictly single-organ support is therefore under-specified for size-against-reference findings, which
may partly explain the inconsistent gains observed for hepatomegaly across encoders, with a clear improvement appearing only for Pillar–0,
whose pretraining likely encodes body-scale context independently.

Several limitations should be acknowledged. First, ORACLE--CT depends on upstream segmentation and can inherit failures from missing, truncated, or inaccurate masks. Although support dilation and fallback handling reduce some practical failure modes, the sensitivity of support-masked pooling to segmentation quality, support size, and encoder-lattice resolution should be quantified more systematically. Second, the label-to-support mapping was manually specified from clinical anatomy and fixed across datasets. This improves interpretability and enables frozen external evaluation, but may be suboptimal for findings with variable, multi-compartment, or weakly localized evidence. Third, labels were examination-level and report-derived, and may contain structured missingness, incomplete co-labeling, and dataset-specific reporting patterns; these issues are especially relevant for rare findings or clinically correlated labels, where models may learn associations that are valid in one dataset but less stable externally. Finally, this study focused on abdominal CT and a single-volume classification setting, and therefore did not address other body regions, lesion-level tasks, multi-series examinations, contrast-phase integration, or longitudinal patient-level modeling.

These limitations suggest several future directions. One direction is to reduce dependence on a separate segmentation pipeline by developing hybrid models that jointly learn anatomical localization and multi-label classification, or that propagate uncertainty from anatomical support maps into the classification head. A second direction is to make support definitions more adaptive: instead of using fixed label-to-support assignments, future models could learn support mixtures, conditionally expand or contract supports, or refine coarse anatomical priors while preserving auditability. Finally, extending anatomy-aware aggregation to multi-series, multi-phase, and longitudinal CT interpretation would better reflect clinical radiology workflows, where decisions often depend on contrast phase, reconstruction type, and comparison with prior examinations. Such extensions will require datasets with more consistently curated cross-series and temporal supervision than is currently available in most large-scale abdominal CT benchmarks.

\section{Conclusion}
\label{sec:conclusion}

We presented ORACLE--CT, an anatomy-aware support-pooling framework for multi-label abdominal CT classification. ORACLE--CT assigns each finding to a predefined anatomical support derived from multi-organ segmentation and restricts evidence aggregation to that support, without changing the encoder architecture or training protocol. Across DINOv3, I3D--ResNet-121, and Pillar--0, the results show that where spatial evidence is aggregated is a major design choice in volumetric CT classification. Support-masked pooling provided the clearest gains for DINOv3 and I3D--ResNet-121, whereas Pillar--0 showed smaller and metric-dependent benefit beyond learned attention, suggesting that radiology-native pretraining may reduce, but not eliminate, the value of explicit anatomical support. Overall, ORACLE--CT provides a practical encoder-agnostic mechanism for making CT classifiers more anatomically aligned, externally robust, and auditable at the evidence-aggregation stage.

\section*{Code, Model and Data Availability}

Code and trained model weights for ORACLE--CT will be made available at public repository upon publication. MERLIN and AMOS are public datasets available from their respective dataset/challenge sources. Duke--Abdomen is a private institutional dataset and cannot be publicly released because of institutional and patient-privacy restrictions.

\section*{Acknowledgements}
Some schematic icons used in Figure~\ref{fig:oracle_overview} and Graphical Abstract were adapted from Servier Medical Art (https://smart.servier.com/), licensed under CC BY 4.0 (https://creativecommons.org/licenses/by/4.0/), and from Flaticon (https://www.flaticon.com/), used according to the respective source licenses.  This work was funded by the Center for Virtual Imaging Trials, NIH grants P41EB028744, R01EB001838,
and R01CA261457.



\FloatBarrier


\bibliographystyle{cas-model2-names}

\bibliography{cas-refs}

\section{Supplementary Material}
\clearpage

\renewcommand{\thetable}{S\arabic{table}}
\renewcommand{\thefigure}{S\arabic{figure}}


\begin{table*}[t]
\centering
\scriptsize
\setlength{\tabcolsep}{2pt}
\renewcommand{\arraystretch}{1.12}
\caption{\textbf{MERLIN full 30-label performance with 95\% bootstrap confidence intervals.}
Values are macro-averaged point estimates with 95\% bootstrap confidence intervals in brackets. Metrics are computed over the full 30-label MERLIN test set. Aggregation modes are abbreviated as GAP, Mask-free attention pooling, and Support-masked pooling. Bold values indicate the best point estimate within each backbone block.}
\label{tab:supp_merlin_full_ci}
\begin{tabular}{llcccc}
\toprule
\textbf{Backbone} & \textbf{Aggregation mode} & \textbf{AUROC} & \textbf{AUPRC} & \textbf{F1} & \textbf{BA} \\
\midrule

\multirow{3}{*}{DINOv3}
& GAP baseline
& 0.838[0.827--0.847]
& 0.638[0.624--0.662]
& 0.636[0.618--0.650]
& 0.730[0.720--0.740] \\

& Mask-free
& 0.847[0.837--0.857]
& 0.662[0.647--0.683]
& 0.643[0.626--0.657]
& 0.726[0.717--0.736] \\

& Support-masked
& \textbf{0.858[0.848--0.867]}
& \textbf{0.676[0.661--0.699]}
& \textbf{0.660[0.643--0.673]}
& \textbf{0.747[0.736--0.756]} \\

\midrule

\multirow{3}{*}{I3D--ResNet-121}
& GAP baseline
& 0.829[0.818--0.839]
& 0.617[0.605--0.640]
& 0.611[0.593--0.624]
& 0.712[0.702--0.722] \\

& Mask-free 
& 0.818[0.807--0.828]
& 0.609[0.595--0.631]
& 0.605[0.587--0.618]
& 0.710[0.700--0.720] \\

& Support-masked 
& \textbf{0.848[0.838--0.857]}
& \textbf{0.659[0.644--0.681]}
& \textbf{0.642[0.626--0.655]}
& \textbf{0.743[0.732--0.753]} \\

\midrule

\multirow{3}{*}{Pillar--0}
& GAP baseline
& 0.800[0.790--0.811]
& 0.598[0.587--0.621]
& 0.597[0.580--0.612]
& 0.692[0.683--0.701] \\

& Mask-free 
& 0.856[0.846--0.865]
& \textbf{0.693[0.680--0.711]}
& 0.657[0.641--0.670]
& \textbf{0.744[0.734--0.753]} \\

& Support-masked 
& \textbf{0.858[0.848--0.867]}
& 0.688[0.674--0.708]
& \textbf{0.660[0.644--0.672]}
& 0.742[0.733--0.752] \\

\bottomrule
\end{tabular}
\end{table*}

\clearpage 




\begin{table*}[t]
\centering
\scriptsize
\setlength{\tabcolsep}{2pt}
\renewcommand{\arraystretch}{1.13}
\caption{\textbf{Harmonized 10-label performance with 95\% bootstrap confidence intervals.}
Values are macro-averaged point estimates with 95\% bootstrap confidence intervals in brackets. Metrics are computed over the 10-label subset shared across MERLIN, Duke--Abdomen, and AMOS. Models were trained on MERLIN; Duke--Abdomen and AMOS were used only for frozen external evaluation. Aggregation modes are abbreviated as GAP, Mask-free attention pooling, and Support-masked pooling. Bold values indicate the best point estimate within each cohort/backbone block.}
\label{tab:supp_harmonized_10_ci}
\begin{tabular}{lllcccc}
\toprule
\textbf{Cohort} & \textbf{Backbone} & \textbf{Aggregation} &
\textbf{AUROC} & \textbf{AUPRC} & \textbf{F1} & \textbf{BA} \\
\midrule

\multirow{9}{*}{MERLIN}
& \multirow{3}{*}{DINOv3}
& GAP
& 0.841[0.825--0.856]
& 0.673[0.648--0.704]
& 0.655[0.629--0.676]
& 0.694[0.678--0.710] \\
& & Mask-free
& 0.846[0.830--0.861]
& 0.696[0.671--0.726]
& 0.670[0.645--0.693]
& 0.696[0.681--0.713] \\
& & Support-masked
& \textbf{0.858[0.843--0.873]}
& \textbf{0.704[0.680--0.736]}
& \textbf{0.681[0.654--0.704]}
& \textbf{0.712[0.694--0.729]} \\

\cmidrule(lr){2-7}
& \multirow{3}{*}{\makecell[l]{I3D--\\ResNet-121}}
& GAP
& 0.824[0.807--0.841]
& 0.641[0.619--0.673]
& 0.628[0.603--0.652]
& 0.680[0.664--0.696] \\
& & Mask-free
& 0.806[0.788--0.824]
& 0.637[0.615--0.667]
& 0.625[0.602--0.647]
& 0.680[0.664--0.695] \\
& & Support-masked
& \textbf{0.847[0.832--0.862]}
& \textbf{0.670[0.648--0.702]}
& \textbf{0.645[0.622--0.667]}
& \textbf{0.704[0.686--0.722]} \\

\cmidrule(lr){2-7}
& \multirow{3}{*}{Pillar--0}
& GAP
& 0.800[0.781--0.819]
& 0.652[0.629--0.684]
& 0.635[0.609--0.657]
& 0.672[0.658--0.687] \\
& & Mask-free
& 0.863[0.848--0.877]
& \textbf{0.746[0.722--0.772]}
& 0.698[0.674--0.718]
& 0.723[0.709--0.737] \\
& & Support-masked
& \textbf{0.885[0.871--0.898]}
& 0.745[0.720--0.772]
& \textbf{0.707[0.684--0.729]}
& \textbf{0.729[0.714--0.744]} \\

\midrule

\multirow{9}{*}{\makecell[l]{Duke--\\Abdomen}}
& \multirow{3}{*}{DINOv3}
& GAP
& 0.802[0.785--0.819]
& 0.628[0.603--0.659]
& 0.601[0.573--0.624]
& 0.662[0.640--0.685] \\
& & Mask-free
& 0.813[0.795--0.830]
& 0.649[0.626--0.678]
& 0.625[0.599--0.647]
& 0.697[0.677--0.719] \\
& & Support-masked
& \textbf{0.835[0.818--0.851]}
& \textbf{0.683[0.659--0.714]}
& \textbf{0.635[0.609--0.657]}
& \textbf{0.701[0.678--0.723]} \\

\cmidrule(lr){2-7}
& \multirow{3}{*}{\makecell[l]{I3D--\\ResNet-121}}
& GAP
& 0.775[0.757--0.793]
& 0.590[0.568--0.623]
& 0.572[0.545--0.595]
& 0.641[0.621--0.662] \\
& & Mask-free
& 0.770[0.748--0.789]
& 0.574[0.550--0.608]
& 0.558[0.531--0.583]
& 0.664[0.642--0.687] \\
& & Support-masked
& \textbf{0.820[0.805--0.835]}
& \textbf{0.646[0.622--0.676]}
& \textbf{0.602[0.576--0.625]}
& \textbf{0.701[0.678--0.725]} \\

\cmidrule(lr){2-7}
& \multirow{3}{*}{Pillar--0}
& GAP
& 0.782[0.764--0.800]
& 0.578[0.556--0.610]
& 0.547[0.515--0.574]
& 0.637[0.621--0.652] \\
& & Mask-free
& 0.849[0.832--0.865]
& \textbf{0.703[0.678--0.736]}
& \textbf{0.651[0.622--0.674]}
& \textbf{0.713[0.692--0.735]} \\
& & Support-masked
& \textbf{0.854[0.838--0.869]}
& 0.703[0.678--0.734]
& 0.610[0.581--0.636]
& 0.693[0.671--0.714] \\

\midrule

\multirow{9}{*}{AMOS}
& \multirow{3}{*}{DINOv3}
& GAP
& 0.742[0.724--0.758]
& 0.312[0.292--0.345]
& 0.297[0.274--0.320]
& 0.603[0.591--0.616] \\
& & Mask-free
& 0.740[0.723--0.755]
& 0.321[0.300--0.355]
& 0.300[0.273--0.324]
& 0.613[0.599--0.626] \\
& & Support-masked
& \textbf{0.762[0.745--0.776]}
& \textbf{0.350[0.327--0.383]}
& \textbf{0.350[0.324--0.371]}
& \textbf{0.643[0.630--0.655]} \\

\cmidrule(lr){2-7}
& \multirow{3}{*}{\makecell[l]{I3D--\\ResNet-121}}
& GAP
& 0.714[0.697--0.730]
& 0.294[0.275--0.325]
& 0.267[0.244--0.290]
& 0.591[0.579--0.602] \\
& & Mask-free
& 0.714[0.698--0.730]
& 0.291[0.271--0.321]
& 0.280[0.254--0.300]
& 0.616[0.604--0.628] \\
& & Support-masked
& \textbf{0.735[0.716--0.752]}
& \textbf{0.327[0.306--0.359]}
& \textbf{0.306[0.282--0.327]}
& \textbf{0.628[0.614--0.642]} \\

\cmidrule(lr){2-7}
& \multirow{3}{*}{Pillar--0}
& GAP
& 0.714[0.696--0.729]
& 0.285[0.268--0.315]
& 0.250[0.225--0.273]
& 0.588[0.577--0.598] \\
& & Mask-free
& 0.757[0.739--0.774]
& 0.393[0.368--0.424]
& \textbf{0.307[0.279--0.331]}
& 0.605[0.593--0.617] \\
& & Support-masked
& \textbf{0.771[0.754--0.788]}
& \textbf{0.402[0.377--0.435]}
& 0.287[0.261--0.311]
& \textbf{0.609[0.597--0.620]} \\

\bottomrule
\end{tabular}
\end{table*}




\clearpage

\begin{table*}[t]
\centering
\scriptsize
\setlength{\tabcolsep}{2pt}
\renewcommand{\arraystretch}{1.12}
\caption{\textbf{Duke--Abdomen shared 27-label performance with 95\% bootstrap confidence intervals.}
Values are macro-averaged point estimates with 95\% bootstrap confidence intervals in brackets. Models were trained on MERLIN, and Duke--Abdomen was used only for frozen external evaluation. Metrics are computed over the 27 labels shared between MERLIN and Duke--Abdomen. Aggregation modes are abbreviated as GAP, Mask-free attention pooling, and Support-masked pooling. Bold values indicate the best point estimate within each backbone block.}
\label{tab:supp_duke_27_ci}
\begin{tabular}{llcccc}
\toprule
\textbf{Backbone} & \textbf{Aggregation} & \textbf{AUROC} & \textbf{AUPRC} & \textbf{F1} & \textbf{BA} \\
\midrule

\multirow{3}{*}{DINOv3}
& GAP
& 0.821[0.804--0.837]
& 0.646[0.631--0.673]
& 0.619[0.598--0.635]
& 0.715[0.700--0.729] \\

& Mask-free
& 0.831[0.816--0.845]
& 0.671[0.655--0.699]
& 0.613[0.592--0.631]
& 0.723[0.710--0.736] \\

& Support-masked
& \textbf{0.848[0.832--0.863]}
& \textbf{0.699[0.684--0.726]}
& \textbf{0.633[0.613--0.650]}
& \textbf{0.737[0.722--0.752]} \\

\midrule

\multirow{3}{*}{I3D--ResNet-121}
& GAP
& 0.791[0.773--0.809]
& 0.614[0.600--0.643]
& 0.596[0.576--0.613]
& 0.688[0.672--0.703] \\

& Mask-free
& 0.793[0.775--0.809]
& 0.608[0.594--0.636]
& 0.589[0.568--0.605]
& 0.704[0.687--0.720] \\

& Support-masked
& \textbf{0.828[0.813--0.842]}
& \textbf{0.662[0.648--0.689]}
& \textbf{0.627[0.607--0.642]}
& \textbf{0.735[0.719--0.749]} \\

\midrule

\multirow{3}{*}{Pillar--0}
& GAP
& 0.775[0.755--0.793]
& 0.594[0.579--0.622]
& 0.550[0.528--0.568]
& 0.668[0.654--0.682] \\

& Mask-free
& 0.828[0.811--0.843]
& 0.682[0.665--0.708]
& \textbf{0.624[0.603--0.642]}
& \textbf{0.721[0.707--0.736]} \\

& Support-masked
& \textbf{0.833[0.818--0.846]}
& \textbf{0.693[0.678--0.718]}
& 0.617[0.597--0.634]
& 0.716[0.703--0.729] \\

\bottomrule
\end{tabular}
\end{table*}

\clearpage 

\begin{figure}[H]
\centering
\includegraphics[width=\textwidth]{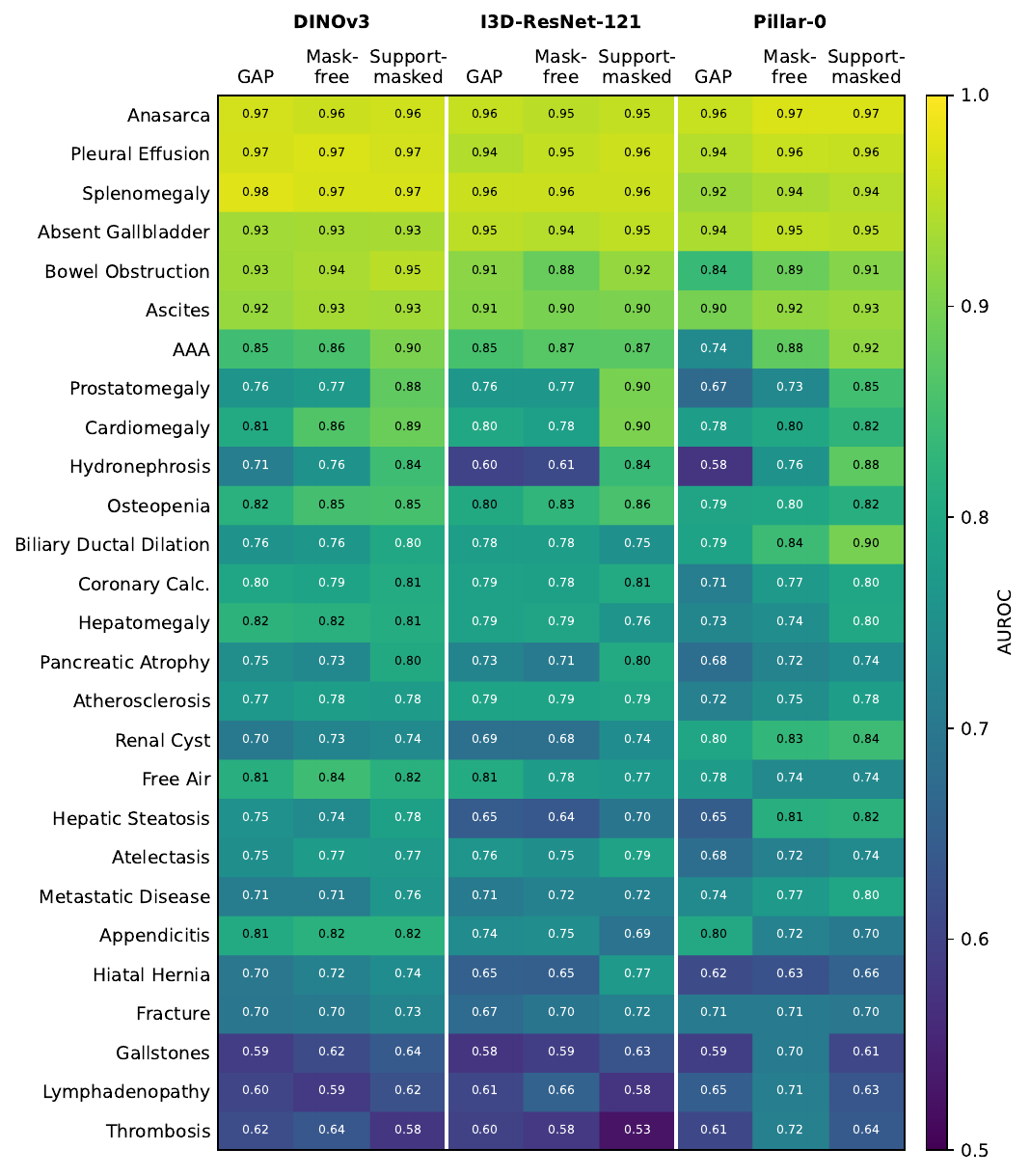}
\caption{\textbf{Absolute per-disease MERLIN-to-Duke transfer on the shared 27-label subset.}
Heatmap of absolute per-finding AUROC on Duke--Abdomen for the 27 findings shared between MERLIN and Duke--Abdomen. Rows correspond to disease labels and columns correspond to aggregation modes across DINOv3, I3D--ResNet-121, and Pillar--0: GAP, mask-free attention pooling, and support-masked pooling. Higher values indicate stronger frozen external discrimination after training on MERLIN. This figure provides the absolute disease-level AUROC values corresponding to the \(\Delta\)AUROC heatmap in the main manuscript.}
\end{figure}

\begin{figure}[H]
\centering
\includegraphics[width=\textwidth]{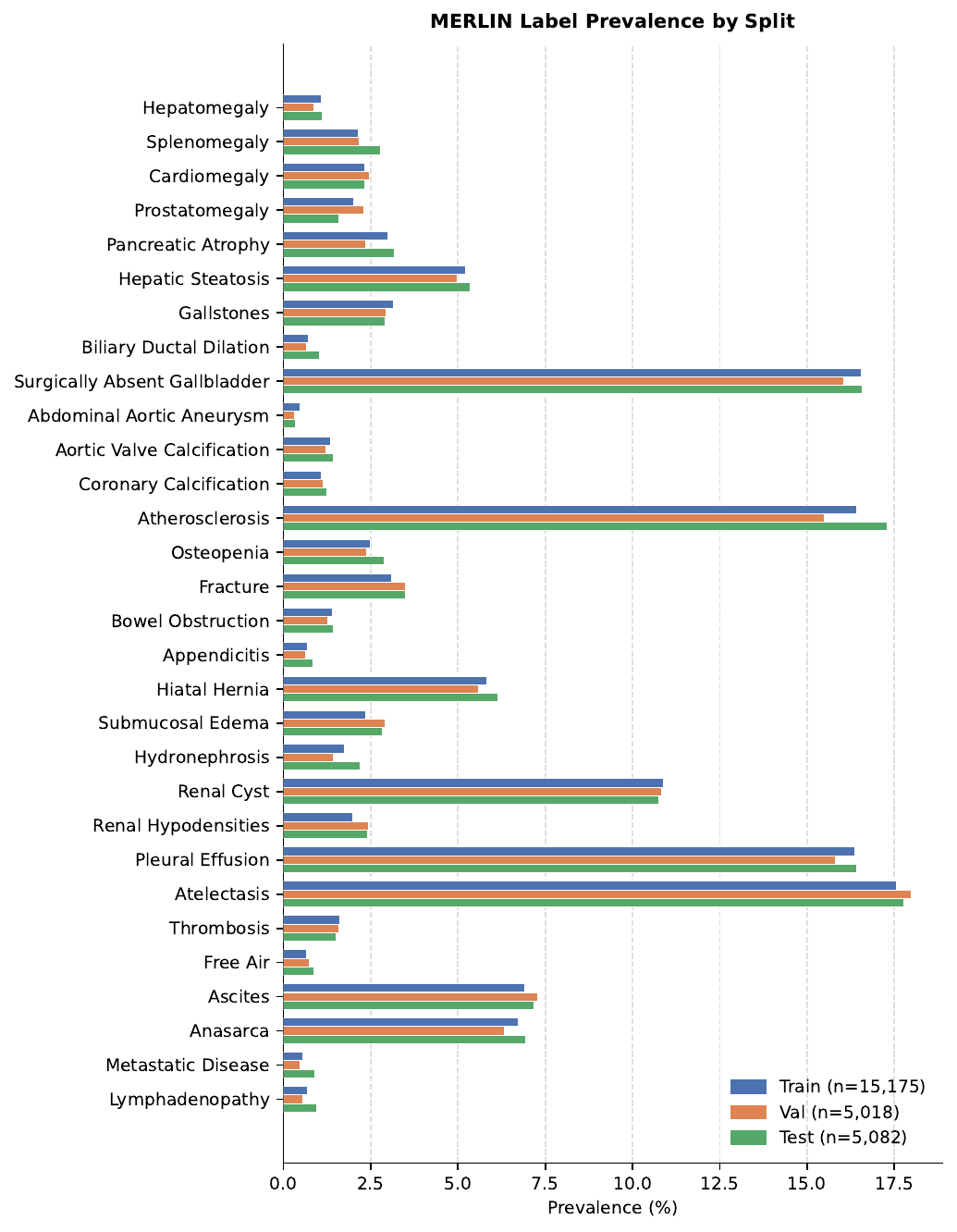}
\caption{\textbf{Label prevalence across MERLIN dataset splits.} Prevalence (percentage of positive cases) for each of the 30 MERLIN findings across the training, validation, and test splits. Prevalence is computed as the fraction of cases labelled positive ($=1$) out of all cases in each split, including those with missing annotations ($=-1$).}
\end{figure}

\end{document}